\documentclass[a4paper,final,10pt,twocolumn]{elsarticle}
\usepackage[vmargin={20mm,20mm},hmargin={15mm,15mm}]{geometry}

\usepackage[utf8]{inputenc}
\usepackage{diagbox}

\usepackage{multirow}

\usepackage{graphicx} 
\usepackage{amsmath}
\usepackage{hyperref}
\usepackage{color}
\usepackage{amssymb} 
\usepackage[normalem]{ulem} 
 \DeclareMathOperator*{\argmin}{argmin}

 \newcommand{\blue}[1]{\textcolor{blue}{#1}}

\usepackage[switch]{lineno}
\modulolinenumbers[1]

 \journal{Pattern Recognition}

\begin{document} 
\begin{frontmatter}

\title{UcoSLAM: Simultaneous Localization and Mapping by Fusion of KeyPoints and Squared Planar Markers}

\author{Rafael~Mu\~noz-Salinas\corref{cor1}\fnref{fn1,fn2}}
\ead{rmsalinas@uco.es}

\author{R. Medina-Carnicer\fnref{fn1}\fnref{fn1,fn2}}
\ead{rmedina@uco.es}

\cortext[cor1]{Corresponding author}
\fntext[fn1]{Computing and Numerical Analysis Department, Edificio Einstein. Campus de Rabanales, C\'ordoba University, 14071, C\'ordoba, Spain, Tlfn:(+34)957212255}
 \fntext[fn2]{Instituto Maim\'onides de Investigaci\'on en Biomedicina (IMIBIC). Avenida Men\'endez Pidal s/n, 14004, C\'ordoba, Spain, Tlfn:(+34)957213861}

 \begin{abstract}
 
This paper proposes a novel approach for Simultaneous Localization and Mapping by fusing natural and artificial landmarks. Most of the SLAM approaches use natural landmarks (such as keypoints). However, they are unstable over time, repetitive in many cases  or insufficient for a robust tracking (e.g. in indoor buildings). On the other hand, other approaches have employed artificial landmarks (such as squared fiducial markers) placed in the environment to help tracking and relocalization. We propose a method that integrates both approaches in order to achieve long-term robust tracking in many scenarios. 

Our method has been compared to the start-of-the-art methods ORB-SLAM2~\cite{orb-slam2} and LDSO~\cite{ldso} in the public dataset  Kitti~\cite{Geiger2012CVPR}, Euroc-MAV~\cite{Burri25012016}, TUM~\cite{sturm12iros} and  SPM~\cite{spm-slam}, obtaining better precision, robustness and speed. Our tests also show that the combination of markers and keypoints achieves better accuracy than each one of them independently.

\end{abstract}

\end{frontmatter}

\begin{figure*}[t!]
\centering
\includegraphics[width=1\textwidth]{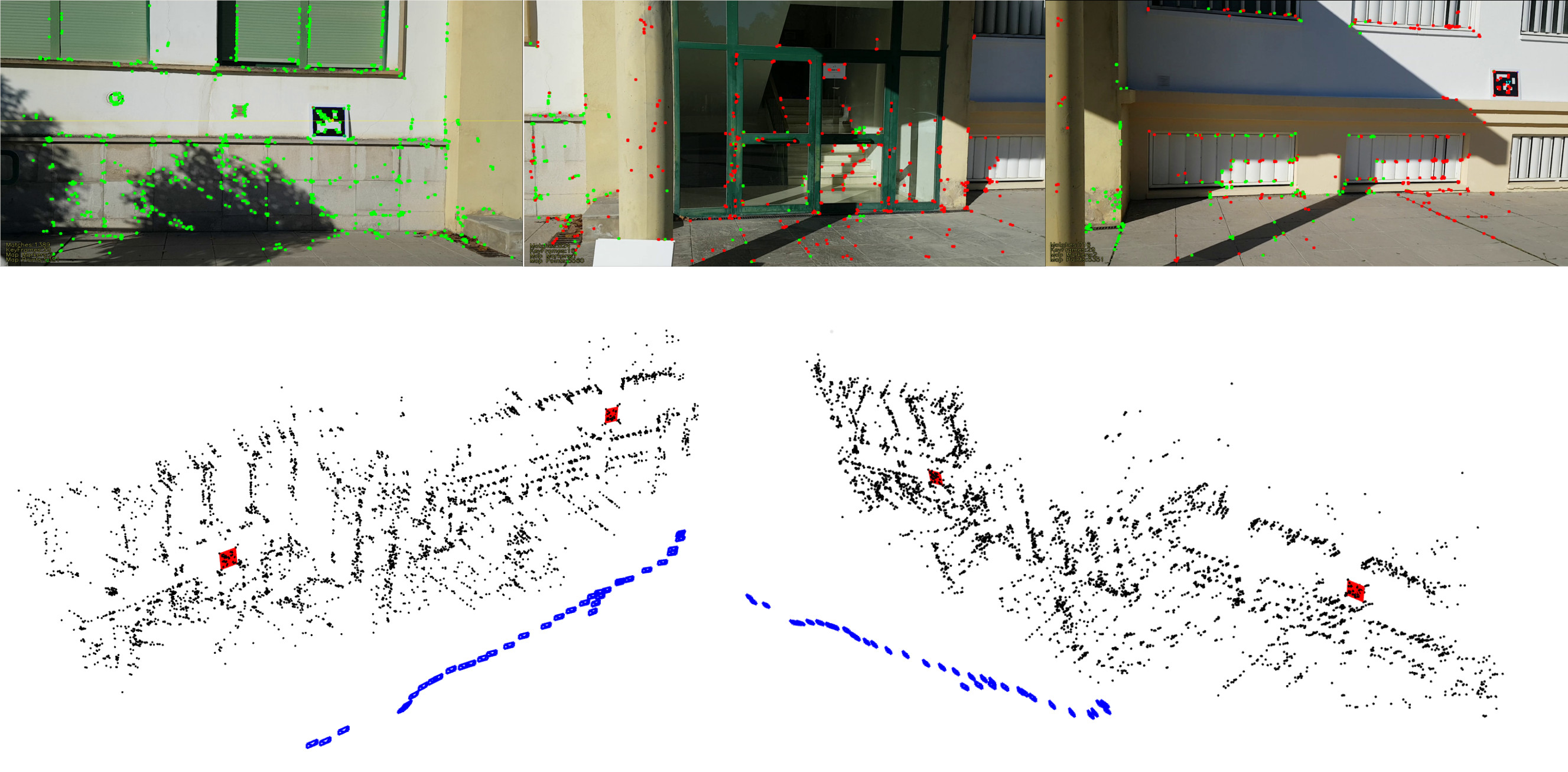}
\caption{Reconstruction example with our method combining keypoints and squared markers. The top row shows some images of the video sequence while the bottom row shows the reconstruction observed from two different viewpoints.  See text for details. }
\label{fig:teaser}
\end{figure*}

\section{Introduction}\label{sec:introduction}
Simultaneous Localization and Mapping is the process of creating a map of the environment while navigating in it \cite{slamreview}. When using monocular cameras, it is a challenging problem for which recent approaches have demonstrated very good results.  The works ORB-SLAM2 \cite{orb-slam2} and LDSO ~\cite{ldso} are probably the state-of-the-art methods in monocular SLAM nowadays. However, these methods suffer from several limitations. First, the scale of the generated maps is unknown, thus making the methods useless in autonomous navigation. Second, they fail in case of pure rotational movements. Third, they require a certain amount of texture, which in some indoor environments is not available (e.g., labs and corridors). Finally, the methods employed for relocalization ({\em bag-of-words} (BoW) \cite{DBOw}),  have a limited performance under viewpoint changes, repetitive patterns and changes over time.
 
Another approach to estimate the pose of a camera consists in using artificial markers placed in the environment. The recent work SPM-SLAM~\cite{spm-slam} solves some of the previously described limitations by using squared fiducial markers (squared patterns with a unique id encoded as a binary matrix) instead of natural features. Markers can be freely placed in the environment and the SPM-SLAM method is able to create a three-dimensional map of the markers. The generated map is in the correct scale and camera localization can be done by observing a single marker. The main drawback of that approach is that it requires to place a lot of markers in the environment in order to build the map. This is because at least two markers must be visible in an image in order create the connection between them. Thus, it can not be applied to large scale problems.
 
This work proposes the system UcoSLAM, a novel SLAM method that fuses keypoints with squared fiducial markers in order to remove many of the limitations of the previous approaches.  First, the scale of the map is automatically determined as soon as a marker is detected in the environment. Second, the proposed method can either use only markers, only keypoints, or a combination of them. Thus, it can operate without problems in markerless environments, and taking advantage of the markers when they are detected. Third, markers help to alleviate the relocalization problem since they remove the visual ambiguity in repetitive environments. In large scale problems, markers can be placed only in strategic locations to help relocalization and reduce drift. In addition, they serve as a tool for long term mapping. While keypoints obtained from an image can change dramatically over time, markers are stable elements that never change. 

Figure~\ref{fig:teaser} shows a reconstruction example generated with our method. The top row shows some images of the video sequence. The colored dots in the images correspond to detected keypoints. The bottom row shows the reconstruction observed from two different viewpoints. Blue elements represent camera frames, black dots 3D points and the red squares the detected markers. 
shows an example reconstructed with our proposed method. As can be observed, many keypoints are extracted in regions that will not last such as shadows. However, the markers can be employed as long-term features for robust tracking and relocalization.

In addition to the proposed UcoSLAM method, we propose a methodology to compare different SLAM methods, considering not only the accuracy in the estimation of the pose but also the percentage of correctly tracking frames.  The proposed method has been compared with the ORB-SLAM2, LDSO and SPM-SLAM using the popular datasets Kitti~\cite{Geiger2012CVPR}, Euroc-MAV~\cite{Burri25012016}, TUM~\cite{sturm12iros} and  SPM~\cite{spm-slam}. The results show that our method is more precise and robust than the other methods. In terms of speed, the proposed method is faster than ORB-SLAM2 and LDSO. Our experiments also show that combining markers and keypoints obtains better precision than markers and keypoints alone.

A final advantage of our work is that it is publicly available for other researchers\footnote{\url{ucoslam.com}}. In contrast to ORB-SLAM2, our implementation allows to load and save the generated maps, which makes it a very useful tool in real applications. In fact, our implementation allows to generate the map in a sequential mode (thus avoiding dropping frames) and saving it for later use. This is specially appropriated in real applications since mapping (a  time-consuming process)  could be generated off-line in a powerful computer, and then the transferred to an autonomous platform with less computing resources for the only purpose of tracking.

The remainder of this paper is structured as follows. Section~\ref{sec::relworks} reviews the most relevant related works. Section~\ref{sec:system_overview} provides an overview of the proposed system that is explained in more detail in Section~\ref{sec::sys-descript}. Finally, Section~\ref{sec:experiments} presents the results and Section~\ref{sec:conclu} draws some conclusions.

\section{Related works}
\label{sec::relworks}

Monocular SLAM aims at solving the simultaneous localization and mapping problem using a single camera exclusively. One of the most influential works is PTAM \cite{PTAM}, presented by Klein and Murray, in which a couple of parallel threads perform tracking and mapping. Using the FAST~\cite{FAST} feature corner detector, they established keypoints matched by patch correlation, which is appropriate for tracking but not for relocalization. However,  their work showed the possibility of splitting the tasks into two different threads, achieving real-time performance.  More recently, Mur-Artal {\it et al.} presented ORB-SLAM~\cite{orb-slam}, a keyframe-based SLAM method using ORB keypoints~\cite{orb} that are employed both for tracking and relocalization using a BoW approach ~\cite{dbow2}. The system computes both the camera position and the structure of the environment by reducing the reprojection error of the keypoints given a set of camera locations known as keyframes. Later, their system is completed in ORB-SLAM2~\cite{orb-slam2} to work not only with monocular cameras but also with stereo ones (including RGBD sensors). As previously mentioned, this system has several drawbacks. First, relocalization is not possible (or reliable) in an environment with repetitive structures such as office-like buildings.  Second, when using monocular cameras, the scale of the generated map is unknown, thus, impeeding SLAM to be used in real-time for navigation purposes.  Third, these methods require a minimum amount of texture in order to extract reliable keypoints. It is simply impossible in some indoor environments with large white walls and ceilings. Finally, many of the keypoints employed for tracking and relocalization correspond to short-term features such as corners in shadows or objects that will move.

Engel and Cremer ({\it et al.}) \cite{engel14eccv,DSO,ldso} employ a different approach to solve the problem. Instead of using keypoints for pose estimation, which is considered an indirect method since the pose is not estimated directly from the image pixels, they use a direct approach consisting in minimizing the photogrammetric error. In their first work (LSD-SLAM), all pixels are subject to minimization. However, in the most recent approaches, they show that only a subset of them are required (sparse approach) ~\cite{DSO}. Their method requires a calibration of the photogrammetric camera properties in order to obtain good results, which may not be possible in some popular datasets. In addition, their first works \cite{engel14eccv,DSO} did not implement relocalization, which is a topic covered in the recent paper LSDO~\cite{ldso}. In any case, the problems above mentioned still present in this SLAM approach.

The above-mentioned approaches assume no intervention in the mapped environment so that only natural features are employed. However, in some use cases, it is possible to alter the environment in order to ease the task. Then, fiducial markers are a very popular approach since they help to improve precision, robustness, and speed. The most simple approach consists in using single points, such as LEDs, retroreflective spheres or planar dots \cite{leds1,leds2}, while other authors have also proposed circular markers encoding a unique identification \cite{circularmarkers1,circularmarkers2} and 2D-barcodes technology   \cite{cybercode,visualcode}. However, the approaches based on squared planar markers are the most popular ones nowadays \cite{Aruco2014,artagPAMI,artoolkit,studierstube,GarridoJurado2015,artagvsartoolkitplus}. A squared planar marker generally consists of an external black border and an internal (most often binary) code to uniquely identify each marker.  Their main advantage is that a single marker provides four correspondence points (its four corners), which are enough to do camera pose estimation.  Despite their advantages, most of the proposed approaches are able to estimate the camera pose only with respect to a single marker, thus making the solution useless in large-scale applications. In fact, mapping and localization from planar markers is a problem scarcely studied in the literature in favor of keypoint-based approaches. 

Davison {\em et al.} presented in \cite{MonoSlam} one of the first methods for monocular SLAM using squared markers for initialization.  Then, Lim and Lee~\cite{5333379} proposed a SLAM method with planar markers using an Extended Kalman-Filter (EKF) to track a robot pose while navigating in an environment. A similar approach is presented in \cite{yamada2009study} for an autonomous blimp. Later, Klopschitz {\em et al.} \cite{Klopschitz07automaticreconstruction} proposed a method where Structure from Motion was first employed to reconstruct the scene, and then the markers were located from the recovered camera poses. Karam {\it et al.} \cite{shaya2012self} presented a  method to create a map of markers as a pose graph where nodes represent markers and edge the relative pose between them. Finally, Neunert {\it et al.} \cite{neunert2015open} proposed to fuse inertial information with fiducial markers using an EKF. More recently, the authors of this paper have proposed Marker Mapper~\cite{markermaper}, a method that allows obtaining a map of markers using only images of the environment. The proposed method deals with ambiguity problem   \cite{Oberkampf1996495,rpp:pami,Collins2014} arising when using planar markers and allows to create very cost tracking system in indoor environments. By freely distributing a set of markers printed on a piece of paper, the method estimates the marker poses from a set of images, given that at least two markers are visible in each image. Afterward, camera localization can be done, in the correct scale with at very high speed~\cite{fastaurco}.  In a posterior work, we proposed SPM-SLAM~\cite{spm-slam}, an online version of the previous work.

This paper proposes UcoSLAM, a novel approach consisting in combining keypoints and squared fiducial markers, having the advantages of both methods. First, the map can be obtained in the correct scale. Second, markers can be used as stable references along time, in contrast to keypoints that may suffer drastic changes even in short periods of time. Third, our method allows robust tracking in repetitive environments such as office-like buildings. Four, our method does not require the markers to operate, unlike our previous works \cite{markermaper,spm-slam}, but it is able to properly use them (if available) in order to improve tracking.


\begin{figure*}[t!]
\centering
\includegraphics[width=1\textwidth]{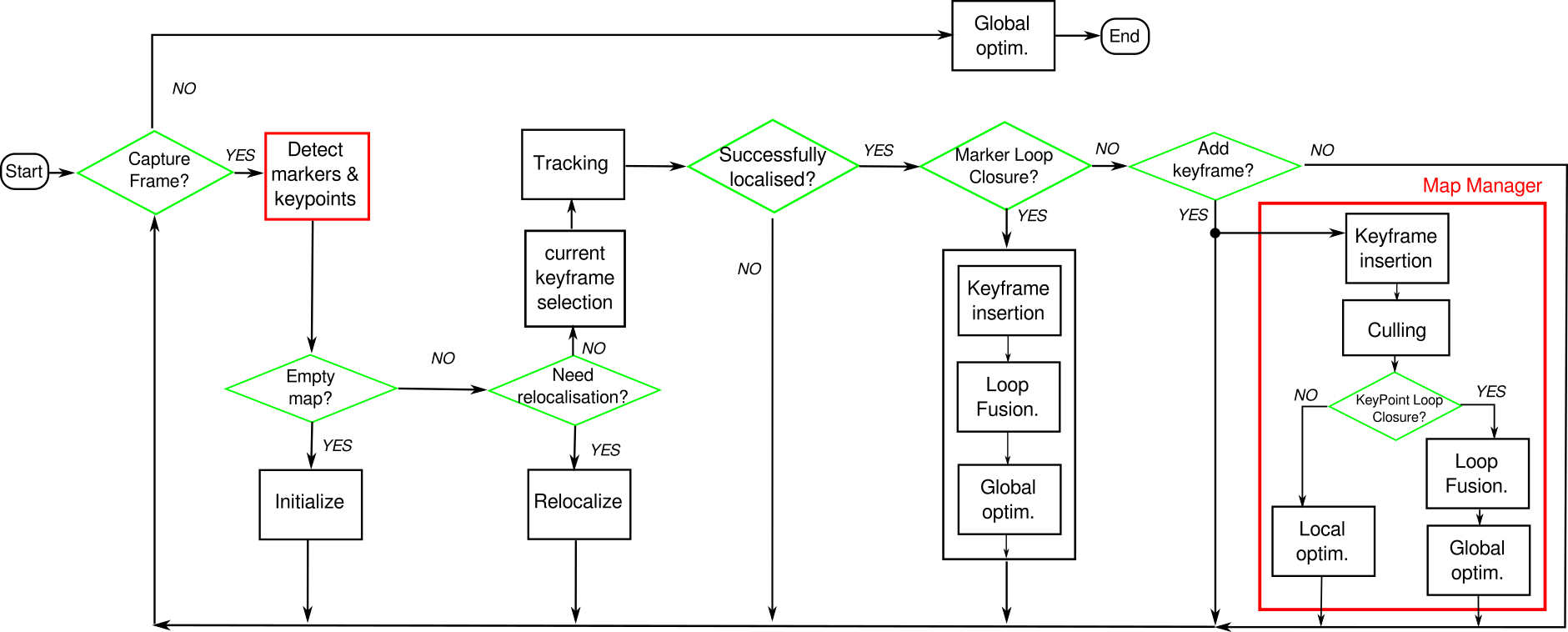}
\caption{Pipeline of the proposed system. }
\label{fig:pipeline}
\end{figure*}

\section{System Overview}
\label{sec:system_overview}
This section provides an overview of the proposed system explaining its most distinctive features.

\subsection{System map and notation employed}

 Let us define the main elements of our system. 
 
A frame captured in time $t$ is represented by 
$$
\mathbf{f}=\{t,\mathrm T,\delta\},
$$
where $\mathrm T\in \mathbb{SE}(3)$ is the pose from which it was acquired, which is the transform moving points from the global reference system (grs) to the camera reference system (crs), and $\delta$ is the set of intrinsic camera parameters, comprised by the focal length, optical centre, and distortion coefficients. The camera intrinsic parameters should be estimated before using the system.

Let be $\mathcal{F}=\{\mathbf {f}\}$ the set of frames. Each frame $\mathbf {f} \in \mathcal{F}$ is first subsampled into an image pyramid using the scale factor $\eta \in (1,\infty)$, and the resulting images processed by a keypoint detector and feature extractor, obtaining several keypoints. Let us define a keypoint $\mathbf{g}$ as the set

 $$
\mathbf{g}=\{l,\mathbf{u},\mathbf{d}\},
$$
where
 $l=\in\mathbb{N}_0$ is the pyramid level, corresponding to the scale factor $\eta^l$ at which it is detected, and $\mathbf u\in\mathbb{R}^2$ its pixel coordinates upsampled to the first pyramid level, $l=0$. In addition, each keypoint has a feature descriptor given by the vector
$$\mathbf{d}=(d_1,\ldots,d_n)~|~d_i\in [0,1].$$

Our system maintains a map of the environment consisting of five sets

$$
\mathcal{W}=\{\mathcal{K},\mathcal{P},\mathcal{M},\mathcal{G},\mathcal{D}\}.
$$
where $\mathcal{K}=\{k\} \subset \mathcal{F}$ is named the keyframes set.

The set $\mathcal{P}=\{\mathbf{p}\}$
is the set of map points
$$
 \mathbf p=\{\mathbf x,\mathbf v, \mathbf{ \hat d}\},
$$
representing points of the environment for which the three-dimensional position $\mathbf x\in\mathbb{R}^3$ has been obtained by triangulation from multiple keyframes. Each map point has a viewing direction $\mathbf v\in \mathbb{R}^3$, and a representative descriptor $\mathbf{ \hat d}$. 

Notice that each map point $\mathbf{p}$ has been observed by several keyframes $k$, and consequently there are several keypoints $\mathbf{g}$ corresponding to point $\mathbf p$.

Also, notice that we select the representative descriptor $\mathbf{ \hat d}$ among the set of keypoints descriptors observing $\mathbf{p}$, as the one minimizing its distance to the rest, i.e.,
if $\mathbf{g}_\mathbf{d}^i, i\in \{1 \cdots n\}$ are the descriptors of keypoints $\mathbf{g}^i$  corresponding to point $\mathbf{p}$:
\begin{equation}
 \mathbf{ \hat d}= \mathrm \{\argmin_{\mathbf{g}_\mathbf{d}^i}\sum_{j \in \{1 \cdots n\}}  dist (\mathbf{g}_\mathbf{d}^j,\mathbf{g}_\mathbf{d}^i)\}.
\end{equation} 
Please notice that the notation employed in this work is such that underscripts are employed to refer to elements of a set. For instance, in the previous equation, $\mathbf{g^i}_\mathbf{d}$ refers to element $\mathbf{d}$ of the set $\mathbf{g}$. 

The viewing  direction $\mathbf{v}$ is a vector indicating the average direction of the keyframes observing the point. If $\mathbf{k}^i$ are the keyframes observing the point $\mathbf{p}$, then
\begin{equation}
\mathbf{v}=\frac{ \sum_i \mathbf{k}^i_{\mathrm{T}_3^{-1}}}
{||\sum_i\mathbf{k}^i_{\mathrm{T}_3^{-1}}||^2_2}
\end{equation}
where $\mathbf{k}^i_{\mathrm{T}_3^{-1}}$ is the third column vector from the inverse transform  matrix $\mathrm{T}^{-1}$ of keyframe $\mathbf{k}^i$, i.e., a vector in the $z$ direction of the keyframe. Thus, $\mathbf{v}$ is the  normalized average of such vectors.

We shall also define
$$
 \Upsilon_p=\{(\mathbf{p},\mathbf{k},\mathbf{g})\},
$$
as the set of tuples registering the observations of map points in the keyframes. In other words, the tuple $(\mathbf{p},\mathbf{k},\mathbf{g})$ is an indication that the map point $\mathbf{p}$ has been observed in the keyframe $\mathbf{k}$ as the keypoint $\mathbf{g}$.

The set $\mathcal{M}=\{ \mathbf m\}$
is the set of markers detected, where each marker

$$
\mathbf m=\{s,\mathrm M,\mathbf{x^1},\mathbf{x^2},\mathbf{x^3},\mathbf{x^4}\},
$$ represents a squared fiducial tag observed in the environment, given by its  length $s$, its pose $\mathrm M \in \mathbb{SE}(3)$, which moves points from the marker reference system (mrs) to the global reference system (grs), and 
$\mathbf{x}^i\in\mathbb{R}^3, \forall i $ the four corners of marker $\mathbf{m}$ defined with respect to its own center as:
\begin{equation}
\begin{tabular}{l} 
$\mathbf{x}^1$=( \begin{tabular}{rrr}~$s$/2,&-$s$/2,&0\end{tabular}), \\
$\mathbf{x}^2$=( \begin{tabular}{rrr}~$s$/2,&~$s$/2,&0\end{tabular}),\\
$\mathbf{x}^3$=( \begin{tabular}{rrr}-$s$/2,&~$s$/2,&0\end{tabular}),\\
$\mathbf{x}^4$=( \begin{tabular}{rrr}-$s$/2,&-$s$/2,&0\end{tabular}).\\ \end{tabular}
\end{equation}

Let us also define a marker observation:
$$
\mathbf {c}_{\mathbf m, \mathbf f}=\{\mathbf {c}_{\mathbf m,\mathbf f}^i\in \mathbb{R}^2, i \in \{1,\ldots,4\}\}$$
as the detection of the marker $\mathbf m$ in the frame $\mathbf{f}$, where the elements $\mathbf{c}_{\mathbf m,\mathbf f}^i$ represents the pixel coordinates of the four marker corners observed in the frame.

As in the previous case, we shall define 
$$
 \Upsilon_m=\{(\mathbf m, \mathbf {k},\mathbf {c}_{\mathbf m,\mathbf f})\},
$$
as the set of marker observations, where each tuple indicates that marker $\mathbf{m}$ is observed in keyframe $\mathbf k$ in the pixel coordinates $\mathbf {c}_{\mathbf m,\mathbf f}$. 

The set $\mathcal{G}$ is a connection graph with the interconnection between keyframes. While vertices represent keyframes,  the weight of edges indicates how strong is the keyframes connection. Each map point shared between two keyframes adds one to the edge weigth between them, while each marker adds four to the edge weight, representing the four corners of the marker. 

Finally, the system also employs a keyframe recognition database $\mathcal{D}$, represented as a BoW with the features of the keyframes in the map as proposed in~\cite{dbow2}. Our implementation\footnote{https://github.com/rmsalinas/fbow}, however is highly optimized using vectorial instructions.

\subsection{Reprojection errors}

Let us define 
\begin{equation}
\Psi(\mathrm T,\mathbf{x},\delta)\in\mathbb{R}^2,
\end{equation}
as a function that given a three-dimentional point $\mathbf{x}\in\mathbb{R}^3$, transform it with the transform $\mathrm T\in \mathbb{SE}(3)$  and then obtains its two-dimentional projection in a camera modelled by the parameters $\delta$. Then, we shall define the reprojection error of a three-dimentional point $\mathbf{x}$ observed at the image coordinates $\mathbf{u}$ as:
\begin{equation}
 e(\mathrm T ,\mathbf{x}, \delta ,\mathbf{u} )= \Psi(\mathrm T , \mathbf{x} ,\delta) - \mathbf{u} .
 \label{eq::point_repj_error_base}
\end{equation}

Using the above defined functions, the global reprojection error of the map points in the keyframes can be defined as:
\begin{equation}
E(\Upsilon_p)= \sum_{(\mathbf{p},\mathbf{k},\mathbf{g})\in \Upsilon_p} 
\mathbf{w_{p}^k}  ~ e(\mathbf{k}_{\mathrm T}, \mathbf{p}_\mathbf x, \mathbf k_\delta, \mathbf{g}_\mathbf u)~ \Omega_{\mathbf{g} }~ e( \mathbf{k}_{\mathrm T}, \mathbf{p}_\mathbf x, \mathbf k_\delta , \mathbf{g}_\mathbf u)^\top.
\label{eq::error_map_points}
\end{equation} 
where $\mathbf{w_{p}^k}$ represents the weight of each map point $\mathbf{p}$ in keyframe $\mathbf{k}$ to calculate the reprojection error.

Also, the parameter $\Omega_{\mathbf g}$ in Eq.~\ref{eq::error_map_points} represents the information matrix employed to modulate the influence of the keypoints errors according to the pyramid level $\mathbf{g}_l$  in which they are detected. It is defined as

\begin{equation}
\Omega_{\mathbf g}=  \frac{1}{\eta^{ \mathbf{g}_l}}  \mathrm I,
\end{equation}
where $\mathrm{I}$ is the $2 \times 2$ identity matrix. In other words, keypoints in the last levels of the pyramid (smaller images), are given less relevance than keypoints in the first pyramid level.

On the other hand, the global reprojection error of the markers of the map is defined as:
\begin{equation}
E(\Upsilon_m)= \sum_{(\mathbf m, \mathbf {k},\mathbf {c}_{\mathbf m,\mathbf f })\in \Upsilon_m} \sum_{i=1}^4 \mathbf{w_m^k}  ||e(\mathbf{k}_{\mathrm T} \cdot \mathbf m_{\mathrm M} , \mathbf{m}_{\mathbf{x}^i}, \mathbf{k}_\delta , \mathbf {c}_{\mathbf m,\mathbf f}^i) ||^2_2,
\label{eq::error_map_markers}
\end{equation}
where $\mathbf{w_{m}^k}$ represents the weight of each marker $\mathbf{m}$ in keyframe $k$ to calculate the reprojection error.

Please notice that the result of the matrix multiplication $\mathbf{k}_{\mathrm T} \cdot \mathbf m_{\mathrm M}$ is a transform moving points from the grs to the crs. 

\subsection{Operational pipeline}
The proposed system operates following the pipeline outlined in Fig.~\ref{fig:pipeline}, which is the most common approach in visual SLAM approaches. The main difference in our case is the combined use of keypoints and squared planar markers. Our system maintains a map of the environment that is created and updated every time a new frame is available. At the beginning, the map is empty and initialization is required. As will be explained below (Sect.~\ref{subsec::initialization}), our map can be initialized using either the homography, the fundamental matrices (using keypoints as in \cite{orb-slam2}), but also using one or several markers (as proposed in \cite{spm-slam}). 

When the map is already initialized, the system goes either into tracking (Sect.~\ref{subsec::tracking}) or relocalisation (Sect.~\ref{subsec::reloc}) mode. If the camera pose was properly determined in the last frame, then, the system tries to estimate the current location using the last one as a starting point. Tracking is performed by jointly optimizing the reprojection errors of keypoints and marker corners. But before tracking, the {\em reference keyframe} $\mathbf{\hat k}$, is selected.  The reference keyframe plays a important role in the system and is selected as the map keyframe with more common matches to the frame analyzed in the previous time instant $\mathbf{f_{t-1}}$.

After successfully tracking, the system must search for loop closures caused by markers. In contrast to keypoint-based loop closure detection, when using markers, this process must be performed in advance to prevent using markers for tracking without proper drift correction (Sect.~\ref{subsec::markerloop}). If loop closure is detected, then, the two sides of the map must be properly connected. Then, the current frame is inserted as a new keyframe, the drift accumulated removed and a global optimization process is applied (Sect.~\ref{subsec::map_optimization}) to correct the whole map. 

If no loop closure is detected, the system must decide whether the current frame must be inserted as a new keyframe. This is a process run in an independent thread named {\em Map Manager}.
If the keyframe is inserted, the map is updated and a culling process is run. Culling allows removing redundant information in order to keep the map manageable in size, following a similar approach to \cite{orb-slam2}. 

Afterward, the system must detect if there is a loop closure by using only keypoints. If not, a local optimization is applied to integrate the new information. As in \cite{orb-slam2}, only the keyframes directly connected to the inserted keyframe are employed in the local optimization. If loop closure is detected, the two side of the loop are fused and a global optimization performed.

If tracking failed in the last frame, then, the system enters in relocalization mode (Sect.~\ref{subsec::reloc}). Relocalisation is done looking first for markers already registered on the map. If no known markers are visible, or they are not reliably detected, relocalization using a BoW approach is attempted.

\subsection{Features and threads}
Our method is not restricted to a unique type of descriptor. It is designed in such a way that any descriptor (binary or real) can be employed. Although our experience indicates that ORB   obtains a good trade-off between speed and robustness, in some scenarios where rotation is not present (e.g. cars, drone mapping), a simpler keypoint descriptor could be employed. Nevertheless, it is the aim of this paper to evaluate the best keypoint descriptor for SLAM.

Our system follows the multithread trend employed in previous works ~\cite{orb-slam2,ldso}. The Fig.~\ref{fig:pipeline} shows as red boxes the components that have been parallelized. The number of threads employed is a parameter in our system that can be adjusted according to properly fit the real number of physical processors available in the platform employed. Nonetheless, our system is also capable of running in sequential mode, which is very interesting when it is desired to obtain the same results in different executions on the same video sequence. 

The task {\em Detect markers} \& {\em keypoints} is one of the most time-consuming operations. When using the keypoints, our system can extract them using up to four threads simultaneously. ORB requires applying the FAST detector on a pyramid of images and our implementation allows to distribute the images among the employed threads in a balanced way. Our experience indicates, however, that no relevant speed-up is obtained using more than two threads. Markers are detected using the method in \cite{fastaurco}, and if desired, the system can do the process in parallel to the detection of the keypoints.
As shown in Fig.~\ref{fig:pipeline}, the process of correcting a loop detected by a marker is done sequentially, as explained later. However, another thread that is continuously running in parallel is the {\em Map Manager}. This is the thread in charge of inserting keyframes, culling map points and keyframes and detecting loop closures by mean of a BoW database. However, when running the system in sequential mode, the Map Manager is not executed as an independent thread, but as a sequential process.



\subsection{Map Serialization}

Our system is concerned with long-term map update and conceived to be used in real applications. For that reason, a very important aspect is the possibility to store the generated map for later uses. 
The main problem with serialization comes from the following requisites. First, it is desired to store the map elements (points, markers, and keyframes) in a data structure that allows random access, such as a vector, so that tracking and mapping could be as fast as possible. Second, as will be explained later, many of the created points will be removed due to lack of temporal consistency in a culling process (Sect. \ref{subsec::kfremoval}). Third, the number of map elements is not known in advance.

A regular vector is not a good option for storing the data for the following reason. Since the number of elements is not known in advance, the vector size will have to be eventually increased, and its data copied from the original to the new vector, which is a time-consuming blocking process, especially as the size of the map increases. In ORBSLAM~\cite{orb-slam2}, the problem is solved by dynamically creating/deleting the map elements, and using their memory references along the code. However, that approach has the problem of making it very complicated to serialize the map. As a consequence, their work does not include any serialization method, and thus, its usage in real applications is limited. Using a list or a tree-indexed structure is not fast enough (as we have already tested).

Our work uses an efficient mixed data structure to solve the above-mentioned problems.  In the beginning, a vector with a fixed size is created. When a new element is created, it is added in the first free vector position, which is employed to identify and access to the element.  Whenever an element is removed, the vector position is marked as free and annotated in a list of free positions so that subsequent insertions are placed in the free positions. When the vector is full, a new vector is created for storing the new elements. With this scheme, accessing to an element requires calculating first the vector in which the element is, and then its position into the vector. The approach allows accessing the elements in constant time without incurring in the overhead of copying data to a resized vector.

\begin{figure}[t]
\centering
 \includegraphics[width=0.40\textwidth]{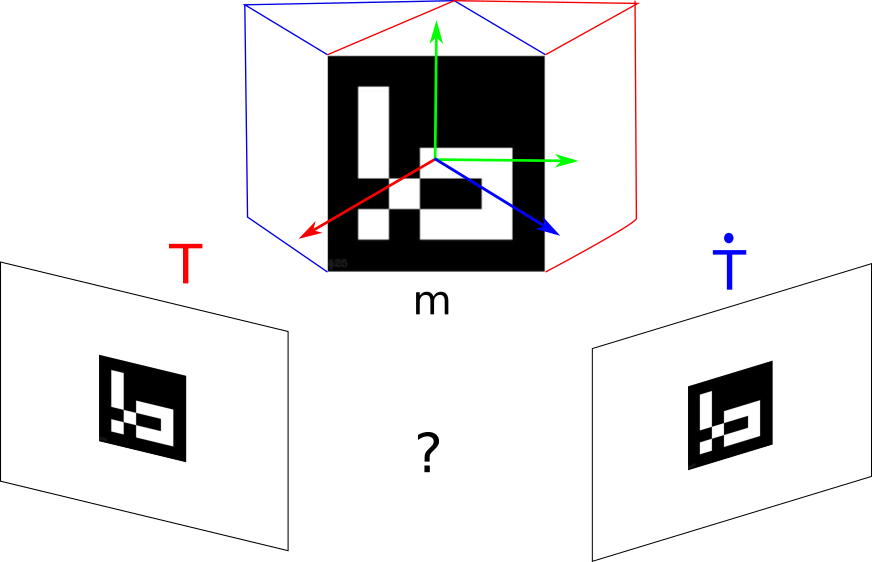}
\caption{Pose ambiguity problem: the same observed projection can be obtained from two different camera poses $T$ and $\dot{T}$. }
\label{fig:ambiguity}
\end{figure}

\subsection{Pose ambiguity in squared planar markers}
\label{sec::ambiguity_problem} 

As already indicated, the pose of a marker wrt a camera can be obtained by using only its four corners. In some cases, however, because of noise in the localization of the corners, there are two possible poses with similar errors. The problem can be observed in Fig~\ref{fig:ambiguity}, where the marker, represented as one side of a cube, could be in two different orientations (red and blue color) thus obtaining almost identical projections from two different camera locations $\mathrm{T}$ and $\dot{\mathrm{T}}$. 
Several authors studied the problem
\cite{Oberkampf1996495,Collins2014,rpp:pami} and proposed specific solutions find the best estimation by a careful analysis of the reprojections. 
In most of the cases, the reprojection error of one of the solutions is smaller than the reprojection error of the other. Then, there is no pose ambiguity and the best solution is the one with the smallest error. But, if the errors in the estimation of the corners become relatively large (e.g. when a marker is far from the camera), the reprojection error of two solutions is very similar and thus impossible to determine the correct one. This work considers the ambiguity problem so as to avoid incorrect pose estimation during initialization and tracking.

\section{Detailed system description}\label{sec::sys-descript}
This section explains in detail the different components involved in the whole process previously outlined.

\subsection{Map Initialization}
\label{subsec::initialization}
The proposed system can initialize the map from a pair of frames using either a keypoints-based approach or a marker-based one. In particular, the initialization process is as follows. We use the first and second frames ($\mathbf f_{\mathbf 0}$ and $\mathbf f_{\mathbf 1}$) and run both initialization methods.  If any of them succeeds,  the map is initialized and we move to the tracking state. If both succeed, priority is given to the marker-based method. If both methods fail,  the operation is repeated using $\mathbf f_{\mathbf 0}$ with the subsequent frames $\mathbf f_{\mathbf 2},\mathbf f_{\mathbf 3},\ldots$. If after $n$ attempts, neither the keyframe-based or marker-based approaches succeed, the role of $\mathbf f_{\mathbf 0}$ is replaced by $\mathbf f_{\mathbf n}$ and so on.

For keypoint-based initialization, the  same approach proposed in \cite{orb-slam2} is applied in our system. It is a robust method that  searches in parallel the homography and essential matrix between the pair of initial frames. While the homography matrix is only valid when the scene is planar, the essential matrix is required otherwise. The problem is to automatically decide which one to use.  The employed method only initializes when it detects enough parallax between the frames, and the solution obtained is robust enough by applying a set of heuristics to select between homography and essential matrices. Because of the restrictions imposed to ensure robustness, initialization from keypoints is sometimes difficult and requires an extra time to be obtained. In addition, maps initialized from keypoints are scale-agnostic, i.e., the real scale of the map can only be recovered up to an scale factor. Nevertheless, our system will automatically rescale the map later if a marker is found along the video sequence.

The marker initialization method employed is the one proposed in \cite{spm-slam}. It can recover the relative pose of the frames and markers, even from ambiguously detected markers, as long as there is enough parallax between the frames. In the presence of ambiguity, each tuple frame-marker generates two solutions, but it is impossible to select the correct one. However, when there are more than one frame observing the markers, the correct solution can be obtained by selecting the solution that minimizes the reprojection errors in all frames.

An advantage of using markers (of known dimensions), is that the initialized map is already in the correct scale and could be directly employed for navigation purposes. In addition, our experience is that employing markers is especially beneficial because it allows a faster initialization than when only keypoints are employed.

\subsection{Tracking}
\label{subsec::tracking}
If the camera pose was successfully estimated in the last frame, the system tries to estimate the current camera pose using the previous one as the starting point. Camera pose estimation consists in minimizing the reprojection error of the set of three-dimensional map points and marker corners observed in the current frame. So, the process is divided into two parts: finding map correspondences, and obtaining the camera pose by minimizing the reprojection errors.

\subsubsection{Finding Map Correspondences}
\label{subsubsec::findmapcorr}
First, matches between the map points observed in the previous frame and the reference keyframe $\mathbf{\hat k}$ are sought, since they are very likely to be present again in the current frame. These matches along with the marker corners observed provide an initial estimation of the current pose. Then, additional matches are found by projecting the map points in the neighbors of the reference keyframe. 

A strict protocol for validating map point matches is followed, as in \cite{orb-slam2}. For each map point, we first check if the angle between its viewing angle $\mathbf{p}_\mathbf{v}$ and the camera direction is smaller than a threshold, otherwise, the point is discarded. Then, we check if the Euclidean distance between the camera and the map point is within the scale invariance region of the keypoint. Afterward, we calculate its projection and if lays out of the image it is discarded. Finally, we compare the point representative descriptor with the keypoints in the image within a search radius (using kd-tree to speed-up computation). We select the two keypoints with the smallest distance within the search radius, and only if the descriptor distance ratio between them is larger than $0.8$, and the smallest descriptor distance is below a minimum threshold $\tau_d$, the match is pre-accepted. Please notice that the value $\tau_d$ is specific to the feature employed. Finally, once all matches have been computed, duplicated assignments are removed: the same keypoint in the current frame could have been assigned to multiple map points. We filter duplicated matches by selecting the one with the smallest distance. Finally, the keypoint matches along with the markers detected are jointly employed to obtain a final refined pose.

An aspect that must be remarked at this point is that not all the visible markers may be employed for tracking in the current frame, for several reasons. As we describe later, a marker is inserted when it is found in a frame. However, its pose may not be estimated at that moment, but delayed until more views of the marker are obtained from different locations. So, visible markers without valid pose are not considered for tracking. Another situation is when a marker with know pose is observed after a long period and its detection implies that a loop closure correction needs to be applied. In that case, the marker cannot be employed for tracking until the accumulated camera pose drift is corrected. Thus, the only markers employed for tracking are these with a valid pose and that have been observed in the neighbors of the reference keyframe.

\subsubsection{Camera Pose Estimation}

Let $$\Upsilon_p^{\mathbf{f}}=\{(\mathbf{p},\mathbf f,\mathbf{g})\}$$  and   $$\Upsilon_p^{\mathbf{f}}=\{ (\mathbf{m},\mathbf f,\mathbf {c}_{\mathbf m,\mathbf f})\}$$ represent the observations of map points  and markers in the frame $\mathbf{f}$, obtained by the procedure explained above. Then, the pose of the frame $\mathbf{f}_{\mathrm T}$, is the unknown to be estimated by minimizing the reprojection errors as: 

\begin{equation}
   \mathbf{f}_{\mathrm T}=\argmin_{\mathrm T} \left( \mathbf{w_p^f} H(\Upsilon_p^{\mathbf{f}},\mathrm T) + \mathbf{w_m^f} H(\Upsilon_m^{\mathbf{f}},\mathrm T) \right),
 \label{eq::bestT}
\end{equation}

where

\begin{equation}
H(\Upsilon_m^{\mathbf{f}},\mathrm T)= \sum_{(\mathbf{m},\mathbf f,\mathbf {c}_{\mathbf m,\mathbf f}) \in \Upsilon_m^{\mathbf{f}} } \sum_{i=1}^4 ||e(\mathrm T \cdot \mathbf m_{\mathrm M} , \mathbf{m}_{\mathbf{x}^i}, \mathbf{f}_\delta , \mathbf {c}_{\mathbf m,\mathbf f}^i ) ||^2_2 ,
\end{equation}
is the sum of the reprojection errors of the markers observed in the current frame, and
\begin{equation}
 H(\Upsilon_p^{\mathbf{f}},\mathrm T)=\\\sum_{(\mathbf{p},\mathbf f,\mathbf{g})\in \Upsilon_p^{\mathbf{f}}} h_\alpha\left( e(\mathrm T,\mathbf{p}_\mathbf{x},\mathbf{f}_\delta,\mathbf{g}_u)~\Omega_\mathbf{g}~e(\mathrm T,\mathbf{p}_\mathbf{x},\mathbf{f}_\delta,\mathbf{g}_u)^\top\right),
 \label{eq::tracking_error_points}
\end{equation}
represents the sum of the reprojection errors of the map points. The Huber function:
\begin{equation}
h_{\alpha }(a)={\begin{cases}{\frac {1}{2}}{a^{2}}&{\text{for }}|a|\leq \alpha \\\alpha (|a|-{\frac {1}{2}}\alpha )&{\text{otherwise}}\end{cases}},
\end{equation}
is employed to decrease the relative importance of outliers in the optimization. The function is quadratic for small values of $a$, and linear for large values, with equal values and slopes of the different sections at the two points where $|a|=\alpha$. 
 
As observed in Eq.~\ref{eq::bestT}, the relative importance of markers and points is weighted by the factors $\mathbf{w_m^f}$ and $\mathbf{w_p^f}$. In general, the number of map points observed in the current frame is much higher than the number of markers. Thus, without weighing their importance, markers would have a negligible importance in the estimation of  $\mathbf{f}_{\mathrm{T}}$. The weights are such that 
$$\mathbf{w_p^f}=1-\mathbf{w_m^f},$$
and
\begin{equation}
\mathbf{w_m^f}= \frac{1}{2} min\left(1,\frac{\mathbf{n_f}}{\tau_m}\right),
\label{eq::wmt}
\end{equation}
where $\mathbf{n_f}$ is the number of valid markers for tracking detected in frame $\mathbf{f}$, and $\tau_m$   is a threshold. The weight $\mathbf{w_m^f}$ is in the range $[0,\frac{1}{2}]$. When there is no valid markers for tracking, $\mathbf{n_f}=0$, the parameter $\mathbf{w_m^f}=0$ and all the importance is given to the map points ($\mathbf{w_p^f}=1$). As, $\mathbf{n_f}\rightarrow \tau_m$, $\mathbf{w_m^f} \rightarrow \frac{1}{2}$, and so the weight of both terms becomes balanced.

\subsection{Keyframe insertion}
\label{subsec::kfadd}

Keyframes are added to the map only if they add new information to the system in order to allow a smooth and reliable tracking. Thus, we follow a set of rules designed to satisfy the need to add both markers and keypoints to the map. The first three rules are aimed at checking the need of adding the keyframe considering the visible markers in the current keyframe. If the keyframe is not added because of these rules, the final rule checks if the frame must be added in order to create new map points. These are the rules employed:
\begin{enumerate}
 \item If the current frame has at least a new marker (a maker not present in the map), the marker and the frame are added to the map. Markers are added the first time they are spotted in a frame even if its pose can not be unambiguosly estimated in it. In that case, the pose of the marker is set as invalid, and it will be later estimated when other keyframes observing the marker are added to the map. This is explained with more detail in Sect.~\ref{subsec::markeraddition}.
 \item If the frame contains at least one marker that has an invalid pose in the map, and the marker pose can be obtained unambiguosly in this frame, then the frame is added and the pose of the marker set in the map.
 \item If the frame contains at least one marker, and the distance from the current frame to the nearest keyframe in the map is larger than a threshold $\tau_b$, the frame is added.
 \item If the number of map points matched in the current frame is below a percentage $\tau_k$ of the total number of map points detected in the reference keyframe, the frame is added. 
 
\end{enumerate}

This last rule is the same as employed in ORB-SLAM2~\cite{orb-slam2}. One a keyframe has been added, either a local or global optimization are run in order to integrate the new information available.

\subsubsection{Initial marker pose estimation}
\label{subsec::markeraddition}

The first time that a new marker is observed in a frame, it is added as keyframe along with the marker. Two cases can occur. First, that the marker is unambiguously detected, and thus its pose $\mathbf{m}_\mathrm{M}$ can be estimated. This is an initial pose subject to subsequent refinements as other keyframes are added to the map (Sect.~\ref{subsec::map_optimization}). The second case is that the marker is ambiguously detected. Then, the marker is added but its pose is set as invalid until it can be later obtained from multiple keyframes. 

Since a new keyframe is added whenever the camera moves away a distance of $\tau_b$ from any other keyframe (rule 3 of KeyFrame insertion in Sect.~\ref{subsec::kfadd}), more frames observing the marker will be added to the map. If any of these keyframes allows an unambiguous detection of the marker pose, then it is annotated. In addition, when there are at least two keyframes observing the marker ambiguously, its pose is obtained as in map initialization (Sect. \ref{subsec::initialization}).

\subsubsection{Map point addition and survival}
\label{subsec::keypointsaddition}
Our system follows a restrictive policy for adding map points, aimed at inserting only robust points. When a new keyframe is added to the map, the system has the opportunity to add new map points, as well as to increase the support for existing ones. For each keypoint we search for possible correspondences among the neighbor keyframes of the reference keyframe. Since the poses of the keyframes are known, the epipolar restrictions are employed to reduce false positives.  Once a point has been added to the map, we apply a survival strategy similar to \cite{orb-slam2}. The basic idea is that the point must be visible at least in two thirds of the following frames until two more keyframes are added to the map. Then, the points become stable and the survival policy is relaxed. Afterward, it only has to be visible in one third of the frames in which it is supposed to be visible. Unlike \cite{orb-slam2}, we always apply the survival policy, even to stable points, in order to allow a long-term mapping reuse.


\subsection{Map optimization}
\label{subsec::map_optimization}

Each time a new frame is added or a loop closure detected, the map needs to be updated in order to integrate the new information, and possibly to remove incorrect map points. The elements subject to optimization are the keyframe poses $\mathbf{k}_\mathrm{T}$, the map point locations $\mathbf{p}_\mathbf{x}$, and the marker poses $\mathbf{m}_\mathrm{M}$.

In essence, the goal of map optimization (see equations (\ref{eq::error_map_points}),(\ref{eq::error_map_markers})) can be enunciated as:

\begin{equation}
 \argmin_{\mathbf{k}_\mathrm{T}\in \mathcal{K},\mathbf{p}_\mathbf{x}\in \mathcal{P},\mathbf{m}_\mathrm{M} \in \mathcal{M} } E(\Upsilon_p)+ E(\Upsilon_m).
 \label{eq:global_opt}
\end{equation}

 Global optimization is an slow process that can be run as a separate thread in the {\em Map Manager}(Fig.~\ref{fig:pipeline}). However, since it is a sparse problem, it can be efficiently optimized using an sparse  version of the Levenberg-Marquardt  algorithm \cite{BundleAdjustment}.
 
 However, optimizing the whole map is only necessary after a loop closure. When a new keyframe is added to the map, only the map keyframes sharing points with it need to be updated, i.e., the neighbors of the new keyframe in the connection graph $\mathcal{G}$. This reduced optimization problem is the so called local optimization.

 Let 
 $$\mathcal{G}_\mathbf{k}\subset\mathcal{K}$$
 the set of keyframes connected to the keyframe $\mathbf{k}$ (including itself), and 

 $$\Upsilon_p(\mathcal{G}_\mathbf{k})\subset \Upsilon_p$$
 and 
 $$\Upsilon_m(\mathcal{G}_\mathbf{k})\subset \Upsilon_m$$
 the subset of observations in those frames. Then, the local optimization can be defined as:

 \begin{equation}
 \argmin_{\mathbf{k}_\mathrm{T}\in \mathcal{G}_\mathbf{k},\mathbf{p}_\mathbf{x}\in \mathcal{P},\mathbf{m}_\mathrm{M} \in \mathcal{M}} E(\Upsilon_p(\mathcal{G}_\mathbf{k}))+ E(\Upsilon_m(\mathcal{G}_\mathbf{k})).
 \label{eq:local_opt}
\end{equation}
\subsection{Keyframe culling}
\label{subsec::kfremoval}
 
Similar to the survival policy employed for points, a keyframe culling strategy is applied to avoid the unlimited growth of   keyframes. Our method, however, must consider that we are dealing both with keypoints and markers. 

The idea is that each marker and map point should be observed by at least three keyframes so as to achieve a good triangulation when doing the optimization processes (Sect.~\ref{subsec::map_optimization}). For each marker $\mathbf{m}$, we select among the set of keyframes in which it is visible $\mathcal{K}_\mathbf{m}$, the three ones with largest distance between them $\mathcal{K}'_\mathbf{m}$. Those frames can not be removed. Then, let  
\begin{equation}
 \mathcal{K}'=\bigcup_{\mathbf{m}\in\mathcal{M}}\mathcal{K}'_\mathbf{m},
\end{equation}
represent the set of keyframes that can not be removed so as to support the markers.

The remaining keyframes are analyzed in order to check if they can be removed without affecting significantly the observed map points. A keyframe is removed if at least $\tau_c\%$ of the keypoints matched to map points are observed in at least another three keyframes in a higher or equal scale of the pyramid (i.e. in image of larger or equal dimensions). In that case, the keyframe can be considered redundant and use the other three keyframes instead.

\subsection{Loop closure detection and correction}
\label{subsec::markerloop}

As the camera moves along the environment, there is an inevitable drift that must be detected and corrected. In our system, the loop closure must differentiate between keypoints and markers.  Loop closure detection using keypoints is a process that can be done as a separated thread into the {\em Map Manager} after adding a new keyframe. To do so, the system keeps a BoW database $\mathcal{D}$ with the keyframes inserted. We follow a similar strategy than \cite{orb-slam2}: the similarity between the recently added keyframe and its neighbors in the graph $\mathcal{G}$ is measured, retaining the lowest score $s_{min}$. Then, the database is queried, obtaining the most similar non-neighbors keyframes, with a similarity higher than $s_{min}$. For each one of these candidates to close the loop,  keypoint correspondences are computed with the inserted keyframe. If the number of correspondences is large enough, we employ the three-dimensional positions of the corresponding map points to find the best rigid transform $\mathrm{T}^l$ between them, using a RANSAC Perspective-n-Point approach. If the number of inliers of the resulting transform is high enough, the loop closure is considered as detected and needs to be corrected.

On the other hand, the detection of the loop closures caused by markers must be done immediately, before processing any other frame. The reason why is that if the loop drift is not corrected before tracking, it can cause camera tracking problems since the uncorrected frame pose can be very different from the pose that would be obtained by the marker. Imagine the typical scenario in which the system is initialized from a marker and the camera moves along the environment until it eventually returns to the initial position. As the camera moves, tracking is based exclusively on keypoints, and the drift increases as usual. When the camera reaches the initial position for the second time and observes the marker again, the estimated camera pose is far from the real one. Then, if the marker was used for camera pose estimation (Eq.~\ref{eq::bestT}), the optimization would have problems to jointly optimize both terms. Therefore,  a marker detected in the current frame is not used in Eq.~\ref{eq::bestT}, unless it has been observed in a neighbor of the reference keyframe $\mathbf{\hat k}$. If this is not the case, then, the detected marker is causing a loop closure.  When the loop closure is detected, two estimations of the keyframe pose are available: one using all available information such as keypoints and the markers not causing the loop detection closure, and another one using the markers causing the loop closure detection. The drift is the difference between these two poses $\mathrm{T}^l$, that must be corrected as explained below.

When a loop closure is detected, either using markers or keypoints, the drift $\mathrm{T}^l$ must be corrected. To do so, we employ the method proposed in \cite{sim3_loopclosure}, that optimizes the Sim(3) keyframe poses. In essence, it uniformly propagates the error along with a path traversing the keyframes from the reference keyframe to the first keyframe in which one of the markers causing the loop closure was detected.

\subsection{Relocalisation}
\label{subsec::reloc}
If tracking failed in the last frame, relocalization must be performed. Relocalisation is first attempted using the markers. If a known marker is unambiguously detected in the current frame,
the pose is estimated. Then, the position is refined by finding map correspondences and reestimating the camera pose (as explained in Sect.~\ref{subsec::tracking}). If the current frame pose cannot be accurately estimated from the markers (because they have been ambiguously detected) the keyframe database $\mathcal{D}$ is employed. It is queried looking for keyframes similar to the current frame and the best ones are analyzed. As for the loop closure detection, keypoints of the current frame are matched with the selected keyframes and the best rigid transform estimated a RANSAC Perspective-n-Point approach. If the number of inliers is high enough relocalization is considered success and the system reenters in tracking mode.


\section{Experiments and results} \label{sec:experiments}

This section explains the experiments conducted to validate the proposed method, and the results obtained. The proposed method, UcoSLAM, has been compared with the state-of-the-art monocular SLAM methods ORB-SLAM2~\cite{orb-slam2} and LDSO~\cite{ldso} using the public datasets Kitti~\cite{Geiger2012CVPR}, Euroc-MAV~\cite{Burri25012016}, TUM~\cite{sturm12iros} and SPM~\cite{spm-slam}.

Along the explanation of our method, a set of parameters have been defined. Table~\ref{tab:params} revisited them indicating the values employed, which provide good results in the test performed.
 
\blue{\begin{table*}[ht!]
\centering
\begin{tabular}{c|c|l}
  Parameter& Default value &   description \\
  \hline
$\tau_d$ & 50 & \parbox{8cm}{Descriptor distance for ORB (Sect.~\ref{subsubsec::findmapcorr})}\\
 & & \\
$\tau_m$ & 5&  \parbox{8cm}{Maximum number of markers  to have a maximum contribution (Eq.~\ref{eq::wmt})}\\
 & & \\
 $\tau_b$ & $0.1$~m & \parbox{8cm}{Minimum baseline distance   between keyframes employed for good   marker triangulation (Sect~\ref{subsec::kfadd}).}\\
 & & \\
$\tau_k$ & $80$\% & \parbox{8cm}{Percentage of matched map points   the refrence keyframe below   a new keyframe is added to the map (Sect~\ref{subsec::kfadd}).}\\
 & & \\
$\tau_c$ & $80$\% & \parbox{8cm}{Minimum percentage of keypoints  in frame observed in   frames for culling
(Sect.~\ref{subsec::kfremoval}).}\\
 \hline
  \end{tabular}
\caption{Main parameters of the proposed system}
  \label{tab:params}
 \end{table*} }

This section is structured as follows. In Sect. ~\ref{sub:comparison_measures} we analyze the most commonly measures used in the literature for SLAM analysis, and propose a new one. Section~\ref{subsec::monoslam_test} compares our method with the state-of-the art methods using only keypoints. Then, in Sect.~\ref{subsec::markersslam_test} we analyze the impact of using markers in tracking. An special use case of a repetitive and low-textured environment is presented in Sect.~\ref{subsec::hallways_test}. We placed markers in the ceiling of a office building and perform a qualitative evaluation of the results. Finally, the computing times of the proposed method is compared to the rest of methods in Sect.~\ref{subsec::computingtimes}.

\subsection{Measures for comparison}
\label{sub:comparison_measures}
This section explains the proposed measure to compare the results of two SLAM methods. Comparing two SLAM methods in a given video sequence with ground truth is a tricky, task since two simultaneous goals are demanded to the SLAM methods. First, it is required to estiate the camera poses as precisely as possible. Second, it is desired to track as many frames as possible. 

Evaluation of the second goal is trivial, but for the first one there are several alternatives. Some authors~\cite{Geiger2012CVPR} propose to   compute the average drift of each frame with respect to its nearby frames. Nonetheless, their  proposal have an important drawback: it reports zero error if the predicted trajectory is in the opposite direction from the real one, as long as it is symmetric. A more informative  measure  of the error between two trajectories that can be computed when the  ground truth is available in all frames is the Absolute Trajectory Error (ATE), which is the translational RMSE after $Sim(3)$ alignment~\cite{engel14eccv}. 

Given the ATE and number of tracked frames of two methods in a sequence, comparing two methods is not a trivial task. Imagine a test sequence of 1000 frames without loop closure. Now, consider a method that is only able to estimate the pose of the camera in the first ten frames  while a second method is able to do a decent tracking in the whole sequence. Because of the low drift in the first ten frames, the total ATE of the first method is small. However, the ATE of the second method in the whole sequence will be larger. Therefore, comparing the final ATE of two methods over the whole sequence is not  fair since they  may have been computed over  a different set of frames. A better approach consists in  comparing the ATEs only in these frames tracked by both methods.

Based on that idea, we propose na ovel methodology to compare SLAM methods in video sequences given the ground-truth. In order to clarify the rest of the explanation, let us denote  $\mathbf{a}$ and  $\mathbf{b}$ two SLAM methods, $E^\mathbf{a}_\mathbf{b}$ the ATE of the method $\mathbf{a}$ given the method $\mathbf{b}$, which is computed as the ATE of the method  $\mathbf{a}$ in the frames tracked by both methods. Similarly, we define $E^\mathbf{b}_\mathbf{a}$ as the ATE of method $\mathbf{b}$ given the method  $\mathbf{a}$. 
Also, let us denote $T_\mathbf{a}$ and $T_\mathbf{b}$ the total number of frames for which the methods $\mathbf{a}$  and $\mathbf{b}$ provide estimations.

Using the ideas explained above, we first propose the score $S^s_\rho(\mathbf{a},\mathbf{b})$ that assigns a value in the range $[0,1]$ to   method $\mathbf{a}$ compared to method $\mathbf{b}$ given the same video sequence $s$.  Higher values indicates that method $\mathbf{a}$ outperforms method $\mathbf{b}$ in sequence $s$. The score is computed using the following set of rules:

\begin {itemize}
 \item  $S^s_\rho(\mathbf{a},\mathbf{b})=1,$ if $E^\mathbf{a}_\mathbf{b}$ is significantly smaller than $E^\mathbf{b}_\mathbf{a}$ and the $T_\mathbf{a}$ is significantly larger than $T_\mathbf{b}$.
\item $S^s_\rho(\mathbf{a},\mathbf{b})=0.5,$ if $E^\mathbf{a}_\mathbf{b}$ is significantly smaller than $E^\mathbf{b}_\mathbf{a}$  but the difference between $T_\mathbf{a}$ and $T_\mathbf{b}$ is not significant.
\item $S^s_\rho(\mathbf{a},\mathbf{b})=0.5,$ if the difference between
$E^\mathbf{a}_\mathbf{b}$ and $E^\mathbf{b}_\mathbf{a}$ is not significant but the $T_\mathbf{a}$ is significantly larger than $T_\mathbf{b}$.
\item $S^s_\rho(\mathbf{a},\mathbf{b})=0,$ otherwise.
\end {itemize}

As can be noticed, the rules give priority to the ATE errors because it is not correct rewarding a method for tracking more frames if the estimated poses have are high error. Thus, the rules only assign a non-zero score to $\mathbf{a}$ if it is, at least, as good as the method $\mathbf{b}$ in terms of ATE. The second aspect considered in the rules is the concept of significance in order to avoid small differences to be considered. 

Please notice that the measure proposed is not commutative. In other words, it is not always true
$$S^s_\rho(\mathbf{a},\mathbf{b}) = S^s_\rho(\mathbf{b},\mathbf{a}).$$ Also, consider that the sum of both is not necessarily one. Actually,
$$S^s_\rho(\mathbf{a},\mathbf{b})+ S^s_\rho(\mathbf{b},\mathbf{a}) \in [0,1].$$

The rules explained above are mathematically formalized as: 
  
  \begin{small}
  \begin{equation}
   S^s_\rho(\mathbf{a},\mathbf{b})=
  \begin{cases} 

  1 & (E^\mathbf{b}_\mathbf{a}-E^\mathbf{a}_\mathbf{b})>  \rho E^\mathbf{a}_\mathbf{b} \wedge (T_\mathbf{a}-T_\mathbf{b})> \rho T_\mathbf{a}  \\
  0.5 &    (E^\mathbf{b}_\mathbf{a}-E^\mathbf{a}_\mathbf{b})>\rho E^\mathbf{a}_\mathbf{b} \wedge |T_\mathbf{a}-T_\mathbf{b}|	\leqslant \hat{\phi}_\rho(T_\mathbf{a},T_\mathbf{b})  \\
  0.5&  |E^\mathbf{b}_\mathbf{a}-E^\mathbf{a}_\mathbf{b}|	\leqslant \phi_\rho(E^\mathbf{b}_\mathbf{a},E^\mathbf{a}_\mathbf{b}) \wedge (T_\mathbf{a}-T_\mathbf{b})  > \rho T_\mathbf{a}    \\
  0 & otherwise\\
  \end{cases}
  \label{eq::ScoreAB}
 \end{equation}
\end{small}

\begin{equation}
 \phi_\rho(x,y)=\rho~min(x,y)
\end{equation}

\begin{equation}
 \hat{\phi}_\rho(x,y)=\rho~max(x,y)
\end{equation}

The equation analyzes if $E^\mathbf{a}_\mathbf{b}$ is greater than   $E^\mathbf{b}_\mathbf{a}$, and if the difference between them is larger than a threshold $\phi_\rho(E^\mathbf{b}_\mathbf{a},E^\mathbf{a}_\mathbf{b})$ controlled by a confidence value $\rho\in(0,1]$.  The threshold avoids infinitesimal differences to be considered as relevant. Consider the case  $E^\mathbf{a}_\mathbf{b}=0.5$ and $E^\mathbf{b}_\mathbf{a}=0.495$, in which the  difference is $0.005$. Using a confidence value $\rho=0.05$ we obtain  $\phi_\rho(E^\mathbf{b}_\mathbf{a},E^\mathbf{a}_\mathbf{b})=0.02475$. So, the differecen between both would not be significant enough to consider the ATE of $\mathbf{a}$ better than the ATE of $\mathbf{b}$. In addition, the score alyzes the difference in the number of tracked frames $(T_\mathbf{a}-T_\mathbf{b})$. As in the previous case, the difference is only considered significant if it is higher than a threshold value $\hat{\phi}_\rho(T_\mathbf{a},T_\mathbf{b})$.

The measure $S^s_\rho(\mathbf{a},\mathbf{b})$ considers only a single video sequence $s$.  In order to obtain a unique score for all the sequences  analyzed $\mathbf{S}$,  we define

\begin{equation}
 \mathbf{S}_\rho(\mathbf{a},\mathbf{b})= \frac{ \sum_{s\in\mathbf{S}} S^s_\rho(\mathbf{a},\mathbf{b})  - S^s_\rho(\mathbf{b},\mathbf{a}) }{ |\mathbf{S}|}
 \label{eq::FinalScore}
\end{equation}

\noindent as a measure in the range $[-1,1]$ comparing both methods. If $\mathbf{S}_\rho(\mathbf{a},\mathbf{b})=1$ it means that method $\mathbf{a}$ scores $1$ in all sequences.  In other words, method $\mathbf{a}$ is better than $\mathbf{b}$ in all sequences. The value $\mathbf{S}_\rho(\mathbf{a},\mathbf{b})=0$ indicates a tie between the methods. And the value $\mathbf{S}_\rho(\mathbf{a},\mathbf{b})=-1$ means that $\mathbf{b}$ scores $1$ in all sequences.

\begin{table*}[p] 
\centering
    \small
    \begin{tabular}{m{2.2cm}||m{2.3cm}||m{.95cm}||m{1.0cm}|m{1.0cm}||m{1.0cm}|m{1.0cm}||m{1.0cm}|m{1.0cm}}
          
          \multirow{2}{1.0cm}{  Dataset }  &
          \multirow{2}{1.0cm}{  Sequence} &
          \multirow{2}{1.0cm}{  Cam} &
          \multicolumn{2}{m{1.5cm}||} {LDSO} & 
          \multicolumn{2}{m{1.5cm}||}{ORB-SLAM2} & 
          \multicolumn{2}{m{1.5cm}}{UcoSLAM}  \\
          
         \cline{4-9} &  & & ATE & \%Trck & ATE & \%Trck & ATE & \%Trck\\
       
       \hline \hline
Euroc-MAV  & V1\_01\_easy           & cam0 & $\infty$        & 0    & $0.082$ & $100.$ & $0.082$ & $100.$\\
Euroc-MAV  & V1\_01\_easy           & cam1 & $\infty$        & 0    & $0.010$ & $100.$ & $0.012$ & $100.$\\
Euroc-MAV  & V1\_02\_medium         & cam0 & $0.000$ & $0.06$ & $0.060$ & $100.$ & $0.061$ & $100.$\\
Euroc-MAV  & V1\_02\_medium         & cam1 & $0.371$ & $1.60$ & $0.028$ & $100.$ & $0.025$ & $100.$\\
Euroc-MAV  & V1\_03\_difficult      & cam0 & $0.000$ & $0.05$ & $0.235$ & $98.4$ & $0.071$ & $95.0$\\
Euroc-MAV  & V1\_03\_difficult      & cam1 & $0.000$ & $0.05$ & $0.178$ & $98.1$ & $0.303$ & $88.8$\\
Euroc-MAV  & V2\_01\_easy           & cam0 & $0.103$ & $100.$ & $0.403$ & $93.5$ & $0.057$ & $100.$\\
Euroc-MAV  & V2\_01\_easy           & cam1 & $0.179$ & $65.9$ & $0.026$ & $99.9$ & $0.019$ & $100.$\\
Euroc-MAV  & V2\_03\_difficult      & cam0 & $\infty$        & 0    & $8.490$ & $90.5$ & $7.930$ & $90.1$\\
Euroc-MAV  & V2\_03\_difficult      & cam1 & $5.460$ & $74.6$ & $4.870$ & $93.8$ & $5.790$ & $87.5$\\
Euroc-MAV  & mh\_01                & cam0 & $\infty$        & 0    & $0.058$ & $100.$ & $0.053$ & $100.$\\
Euroc-MAV  & mh\_01                & cam1 & $0.095$ & $100.$ & $0.061$ & $100.$ & $0.055$ & $100.$\\
Euroc-MAV  & mh\_02                & cam0 & $0.092$ & $100.$ & $0.059$ & $100.$ & $0.057$ & $100.$\\
Euroc-MAV  & mh\_02                & cam1 & $1.670$ & $100.$ & $0.052$ & $100.$ & $0.055$ & $100.$\\
Euroc-MAV  & mh\_03                & cam0 & $0.000$ & $0.04$ & $0.032$ & $100.$ & $0.035$ & $100.$\\
Euroc-MAV  & mh\_03                & cam1 & $0.000$ & $0.04$ & $0.027$ & $100.$ & $0.031$ & $100.$\\
Euroc-MAV  & mh\_04                & cam0 & $0.244$ & $100.$ & $0.041$ & $100.$ & $0.052$ & $100.$\\
Euroc-MAV  & mh\_04                & cam1 & $5.820$ & $100.$ & $0.141$ & $100.$ & $0.092$ & $100.$\\
Euroc-MAV  & mh\_05                & cam0 & $0.116$ & $100.$ & $0.043$ & $100.$ & $0.051$ & $100.$\\
Euroc-MAV  & mh\_05                & cam1 & $0.300$ & $100.$ & $0.049$ & $100.$ & $0.054$ & $100.$\\
Kitti      & 00                   & cam0 & $\infty$        & 0    & $5.930$ & $100.$ & $4.630$ & $100.$\\
Kitti      & 00                   & cam1 & $119.0$ & $100.$ & $5.660$ & $100.$ & $4.000$ & $100.$\\
Kitti      & 01                   & cam0 & $6.180$ & $100.$ & $423.0$ & $97.7$ & $362.0$ & $100.$\\
Kitti      & 01                   & cam1 & $7.800$ & $100.$ & $154.0$ & $59.3$ & $346.0$ & $100.$\\
Kitti      & 02                   & cam0 & $\infty$        & 0    & $18.40$ & $100.$ & $31.10$ & $100.$\\
Kitti      & 02                   & cam1 & $99.80$ & $55.3$ & $11.40$ & $100.$ & $28.50$ & $100.$\\
Kitti      & 03                   & cam0 & $\infty$        & 0    & $1.800$ & $12.2$ & $8.400$ & $100.$\\
Kitti      & 03                   & cam1 & $\infty$        & 0    & $4.140$ & $100.$ & $626.0$ & $51.0$\\
Kitti      & 04                   & cam0 & $\infty$        & 0    & $0.623$ & $100.$ & $0.354$ & $100.$\\
Kitti      & 04                   & cam1 & $\infty$        & 0    & $0.847$ & $100.$ & $0.454$ & $100.$\\
Kitti      & 05                   & cam0 & $\infty$        & 0    & $4.890$ & $98.6$ & $2.670$ & $100.$\\
Kitti      & 05                   & cam1 & $\infty$        & 0    & $6.040$ & $100.$ & $3.540$ & $100.$\\
Kitti      & 06                   & cam0 & $\infty$        & 0    & $15.80$ & $98.7$ & $44.00$ & $99.9$\\
Kitti      & 06                   & cam1 & $\infty$        & 0    & $10.60$ & $96.9$ & $16.00$ & $99.7$\\
Kitti      & 07                   & cam0 & $\infty$        & 0    & $3.370$ & $96.7$ & $2.150$ & $100.$\\
Kitti      & 07                   & cam1 & $\infty$        & 0    & $4.680$ & $98.9$ & $2.230$ & $100.$\\
Kitti      & 08                   & cam0 & $\infty$        & 0    & $42.10$ & $100.$ & $43.40$ & $100.$\\
Kitti      & 08                   & cam1 & $\infty$        & 0    & $33.10$ & $100.$ & $42.70$ & $100.$\\
Kitti      & 09                   & cam0 & $\infty$        & 0    & $24.00$ & $86.8$ & $7.160$ & $100.$\\
Kitti      & 09                   & cam1 & $\infty$        & 0    & $7.580$ & $100.$ & $7.150$ & $100.$\\
SPM        & video1               & cam0 & $2.390$ & $96.7$ & $0.102$ & $67.1$ & $1.430$ & $88.5$\\
SPM        & video2               & cam0 & $1.810$ & $99.8$ & $0.048$ & $99.8$ & $0.054$ & $99.8$\\
SPM        & video3               & cam0 & $0.502$ & $41.0$ & $0.193$ & $99.8$ & $0.098$ & $99.8$\\
SPM        & video4               & cam0 & $1.380$ & $52.1$ & $1.470$ & $99.7$ & $0.011$ & $99.8$\\
SPM        & video5               & cam0 & $0.000$ & $0.04$ & $0.255$ & $64.6$ & $0.023$ & $99.2$\\
SPM        & video6               & cam0 & $0.000$ & $0.04$ & $0.652$ & $29.5$ & $1.460$ & $94.7$\\
SPM        & video7               & cam0 & $0.138$ & $7.50$ & $1.070$ & $99.9$ & $1.760$ & $100.$\\
SPM        & video8               & cam0 & $\infty$        & 0    & $0.064$ & $7.63$ & $0.049$ & $99.8$\\
TUM        & fr1\_desk2            & cam0 & $0.000$ & $0.15$ & $0.150$ & $7.82$ & $1.190$ & $82.9$\\
TUM        & fr1\_desk             & cam0 & $0.930$ & $100.$ & $0.012$ & $100.$ & $0.012$ & $100.$\\
TUM        & fr1\_floor            & cam0 & $1.300$ & $100.$ & $0.024$ & $98.6$ & $0.197$ & $96.9$\\
TUM        & fr1\_xyz              & cam0 & $0.076$ & $99.8$ & $0.008$ & $99.8$ & $0.008$ & $99.8$\\
TUM        & fr2\_360\_kidnap       & cam0 & $0.000$ & $0.06$ & $0.204$ & $66.2$ & $\infty$        & 0   \\
TUM        & fr2\_desk\_person      & cam0 & $0.000$ & $0.02$ & $0.000$ & $0.24$ & $0.002$ & $0.36$\\
TUM        & fr2\_desk             & cam0 & $0.148$ & $37.4$ & $0.081$ & $99.9$ & $0.080$ & $99.9$\\
TUM        & fr2\_xyz              & cam0 & $0.013$ & $99.9$ & $0.004$ & $99.9$ & $0.004$ & $99.9$\\
TUM        & fr3\_long\_office      & cam0 & $1.280$ & $99.9$ & $0.023$ & $99.9$ & $0.019$ & $99.9$\\
TUM        & fr3\_nstr\_tex\_near    & cam0 & $0.924$ & $74.4$ & $0.052$ & $99.9$ & $0.027$ & $99.9$\\

\end{tabular}
    \caption{Results of the different methods tested on the datasets Euroc-MAV~\cite{Burri25012016},  Kitti~\cite{Geiger2012CVPR}, SPM~\cite{spm-slam} and TUM~\cite{sturm12iros}. The symbol $\infty$ correspond to sequences in which the method was not able to initialize.}

    \label{tab:monocular_complete}
\end{table*}

\subsection{Monocular SLAM}
\label{subsec::monoslam_test}
This section compares the proposed method with another two state-of-the-art monocular SLAM methods, namely ORB-SLAM2~\cite{orb-slam2} and LDSO~\cite{ldso} in the popular datasets Kitti~\cite{Geiger2012CVPR}, Euroc-MAV~\cite{Burri25012016}, TUM~\cite{sturm12iros} and SPM~\cite{spm-slam}, all having ground truth information. The Kitti dataset was recorded with a car, mounting a stereo camera on top.  The Euroc-MAV dataset was recorded using a stereo camera mounted on a small drone. The TUM dataset was recorded with a RGBD camera. Finally, the SPM dataset was recorded in a room with markers placed in the walls and the ceiling. Despite the markers present in the sequence, in this experiment we disabled the marker detection of our system so that the comparison is restricted only to keypoints.

For our experiments, only monocular information was employed. As a consequence, we have a total $20$ video sequences for Kitti (corresponding to the left and right cameras), another $20$ video sequences for Euroc-MAV, we selected $10$ sequences from the TUM dataset and the $8$ sequences of the SPM dataset. Since each dataset has different formats for the images and the ground truth, we have created a normalized meta-dataset on which the tests have been performed. Our meta-dataset stores the sequences as mp4 video files compressed using the H264 algorithm. The meta-daset is publicly available for download\footnote{\url{ucoslam.com}}.

The following methodology has been employed in order to analyze the video sequences. The sequence has been first processed to obtain the map. Then, using the generated map, the sequence is processed again and the estimated poses stored. This allows evaluating the methods after correcting loop closures. 

Both LDSO  and our method allows running the system sequentially. In other words, a frame is not processed until the previous one has been thoroughly analyzed. Thus, two runs of the methods using the same sequence produce the same results. The code LDSO code to do out experiments is available online\footnote{\url{https://github.com/rmsalinas/LDSO}}. Since ORB-SLAM2 implementation runs multiple threads in parallel, obtaining repeatable results is not possible. To avoid that problem, we have modified the original code so that it runs sequentially, without dropping frames, and obtaining identical results in different runs on the same sequence. To do so we added mutexes in critical parts of the system. The modified version of ORB-SLAM2 employed is publicly available \footnote{\url{https://github.com/rmsalinas/ORB_SLAM2}}.

The results of the different methods tested are shown in Table~\ref{tab:monocular_complete}. The table shows for each sequence the results of each one of the methods tested. The column {\it Cam} is employed to distinguish the camera employed in the dataset with stereo cameras, being {\it cam0} the left camera and {\it cam1} the right one.  The symbol $\infty$   in the table correspond to sequences in which the method was not able to initialize, so the number of tracked frames is $0$. Since the raw data in the Table does not allows a comparison of the different methods we apply the proposed comparison methodology (Eq.~\ref{eq::FinalScore}) for every pair-wise combination of methods, and using different confidence values. The results are  shown in Table~\ref{tab:mono_score}.

 \begin{table}[ht!]
\centering
\begin{tabular}{l|c|c|r}
  $\rho$&   $\mathbf{a}$ &   $\mathbf{b}$ &  $\mathbf{S}_\rho(\mathbf{a},\mathbf{b})$ \\
  \hline
  $0.01$ & LDSO  & ORB-SLAM2 & -0.34 \\
  $0.01$ & LDSO  & UcoSLAM & -0.37\\
  $0.01$ & ORB-SLAM2  & UcoSLAM & -0.14\\
  \hline
  $0.05$ & LDSO  & ORB-SLAM2 & -0.34 \\
  $0.05$ & LDSO  & UcoSLAM & -0.37\\
  $0.05$ & ORB-SLAM2  & UcoSLAM & -0.12\\
  \hline
  $0.1$ & LDSO  & ORB-SLAM2 & -0.30 \\
  $0.1$ & LDSO  & UcoSLAM &  -0.40\\
  $0.1$ & ORB-SLAM2  & UcoSLAM &  -0.12\\
  \hline
  $0.25$ & LDSO  & ORB-SLAM2 & -0.30 \\
  $0.25$ & LDSO  & UcoSLAM & -0.37\\
  $0.25$ & ORB-SLAM2  & UcoSLAM & -0.10\\
  \hline
  \end{tabular}
\caption{Comparison between the three methods tested. LDSO is the worst method. UcoSLAM is slightly better than ORB-SLAM2.}
  \label{tab:mono_score}
 \end{table} 
 
\begin{figure*}[t!]
\centering
\includegraphics[width=1\textwidth]{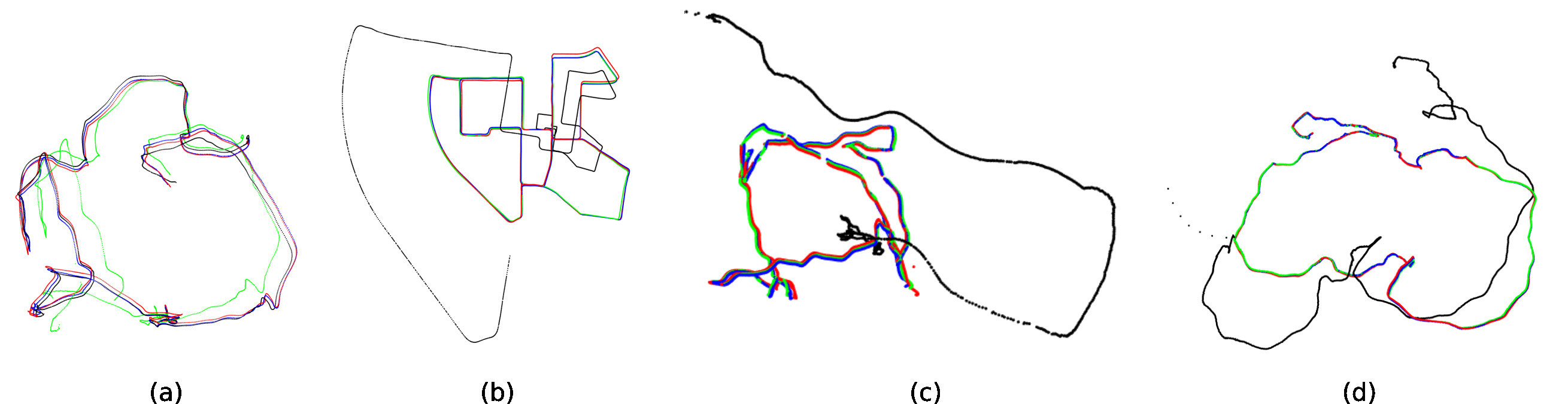}
\caption{ Trajectories of the methods tested on some of the sequences. Red: ground-truth. Blue:UcoSLAM. Green:ORB-SLAM2. Black:LDSO. (a) Euroc-MAV,  V2\_01\_easy, cam0. (b) Kitti, 00, cam1. (c) SPM, video2, cam0. (d) TUM, fr3\_long\_office, cam0}
\label{fig:monoresults}
\end{figure*}

As can be observed, the method LDSO is the worst one, loosing in almost all comparisons. When comparing ORB-SLAM2 and UcoSLAM, our implementation seems to be slightly better. Figure~\ref{fig:monoresults} shows the results of the three methods tested on four the sequence employed.  

It can  then be said, that the proposed implementation for monocular SLAM using keypoints is better than the other two methods tested. 

\subsection{SPM dataset using markers}
\label{subsec::markersslam_test}

In this section we analyze the results of our method when markers are employed for tracking in the SPM dataset. The SPM dataset is comprised by  eight video sequences recorded in our lab, where an Optitrack motion capture system was employed to estimate the pose of a PtGrey FLEA3 camera capturing images of $1920\times1080$ pixels at a frame rate of $60$~Hz. The room had squared fiducial markers placed in the walls and ceiling. Five of the sequences point towards the walls while in the other three sequences the camera is always facing to the ceiling.

For this experiment, we compare our UcoSLAM method using three different variants. First, using only keypoints (as in the previous experiment). Second, using only markers, which is equivalent to our recent work ~\cite{spm-slam}. Finally, using both keypoints and markers. The raw results are shown in Table~\ref{tab:spmdataset} using the names {\it UcoSLAM(kp),UcoSLAM(m),UcoSLAM(kp+m)} for the three versions: with keypoints, with markers, and with keypoints and markers. 
    
As in the previous experiment, the score $\mathbf{S}_\rho(\mathbf{a},\mathbf{b})$ (Eq.~\ref{eq::FinalScore}) has been employed to do a pair-wise comparison of the methods for different confidence values. The results are presented in Table~\ref{tab:spm_score}. As can be observed, in the SPM dataset, the use of markers obtains better results than keypoints,  and the combined use of keypoints and markers obtains the best results. 

\begin{table*}[ht!] 
\centering
    \small
    \begin{tabular}{m{2.2cm}||m{2.3cm}||m{.95cm}||m{1.0cm}|m{1.0cm}||m{1.0cm}|m{1.0cm}||m{1.0cm}|m{1.0cm}}
          
          \multirow{2}{1.0cm}{  Dataset }  &
          \multirow{2}{1.0cm}{  Sequence} &
          \multirow{2}{1.0cm}{  Cam} &
          \multicolumn{2}{m{1.5cm}||} {UcoSLAM(kp)} & 
          \multicolumn{2}{m{1.5cm}||}{UcoSLAM(m)} & 
          \multicolumn{2}{m{1.5cm}}{UcoSLAM(kp+m)}  \\
          
         \cline{4-9} &  & & ATE & \%Trck & ATE & \%Trck & ATE & \%Trck\\
       
       \hline \hline
SPM        & video1               & cam0 & $0.601$ & $64.5$ & $0.065$ & $98.4$ & $0.057$ & $100.$\\
SPM        & video2               & cam0 & $0.057$ & $99.8$ & $0.048$ & $98.4$ & $0.051$ & $99.8$\\
SPM        & video3               & cam0 & $0.108$ & $99.8$ & $0.054$ & $98.1$ & $0.053$ & $99.8$\\
SPM        & video4               & cam0 & $0.010$ & $99.8$ & $0.013$ & $99.0$ & $0.008$ & $99.8$\\
SPM        & video5               & cam0 & $0.015$ & $99.5$ & $0.014$ & $98.8$ & $0.011$ & $99.7$\\
SPM        & video6               & cam0 & $0.687$ & $50.0$ & $0.014$ & $98.6$ & $0.011$ & $99.7$\\
SPM        & video7               & cam0 & $1.280$ & $100.$ & $0.053$ & $98.7$ & $0.050$ & $100.$\\
SPM        & video8               & cam0 & $0.052$ & $99.8$ & $0.062$ & $99.2$ & $0.071$ & $99.8$\\
       \end{tabular}
    \caption{ Results of the different UcoSLAM versions in the SPM dataset. See text for details. }
    \label{tab:spmdataset}
\end{table*}
 
 \begin{table}[ht!]
\centering
 \begin{tabular}{l|c|c|r}
  $\rho$&   $\mathbf{a}$ &   $\mathbf{b}$ &  $\mathbf{S}_\rho(\mathbf{a},\mathbf{b})$ \\
  \hline
  $0.01$ & UcoSLAM(kp)  & UcoSLAM(m) & -0.375 \\
  $0.01$ & UcoSLAM(kp)  & UcoSLAM(m+kp) & -0.500\\
  $0.01$ & UcoSLAM(m)  & UcoSLAM(m+kp) & -0.625\\
  \hline
  $0.05$ & UcoSLAM(kp)  & UcoSLAM(m) & -0.375 \\
  $0.05$ & UcoSLAM(kp)  & UcoSLAM(m+kp)& -0.500\\
  $0.05$ & UcoSLAM(m)  & UcoSLAM(m+kp)& -0.312\\
  \hline
  $0.1$ & UcoSLAM(kp)  & UcoSLAM(m) & -0.312 \\
  $0.1$ & UcoSLAM(kp)  & UcoSLAM(m+kp) &  -0.500\\
  $0.1$ & UcoSLAM(m)  & UcoSLAM(m+kp) &  -0.187\\
  \hline
  $0.25$ & UcoSLAM(kp)  & UcoSLAM(m) & -0.375\\
  $0.25$ & UcoSLAM(kp)  & UcoSLAM(m+kp) & -0.375\\
  $0.25$ & UcoSLAM(m)  & UcoSLAM(m+kp) & -0.187\\
  \hline
  \end{tabular}
\caption{Analysis of the impact of combining marker and keypoints in the SPM dataset. The proposed combination of keypoints and markers improves the other approaches.}
\label{tab:spm_score}
\end{table} 

\subsection{Large scale mapping in low-textured and repetitive environments}
\label{subsec::hallways_test}

\begin{figure*}[t!]
\centering
\includegraphics[width=1\textwidth]{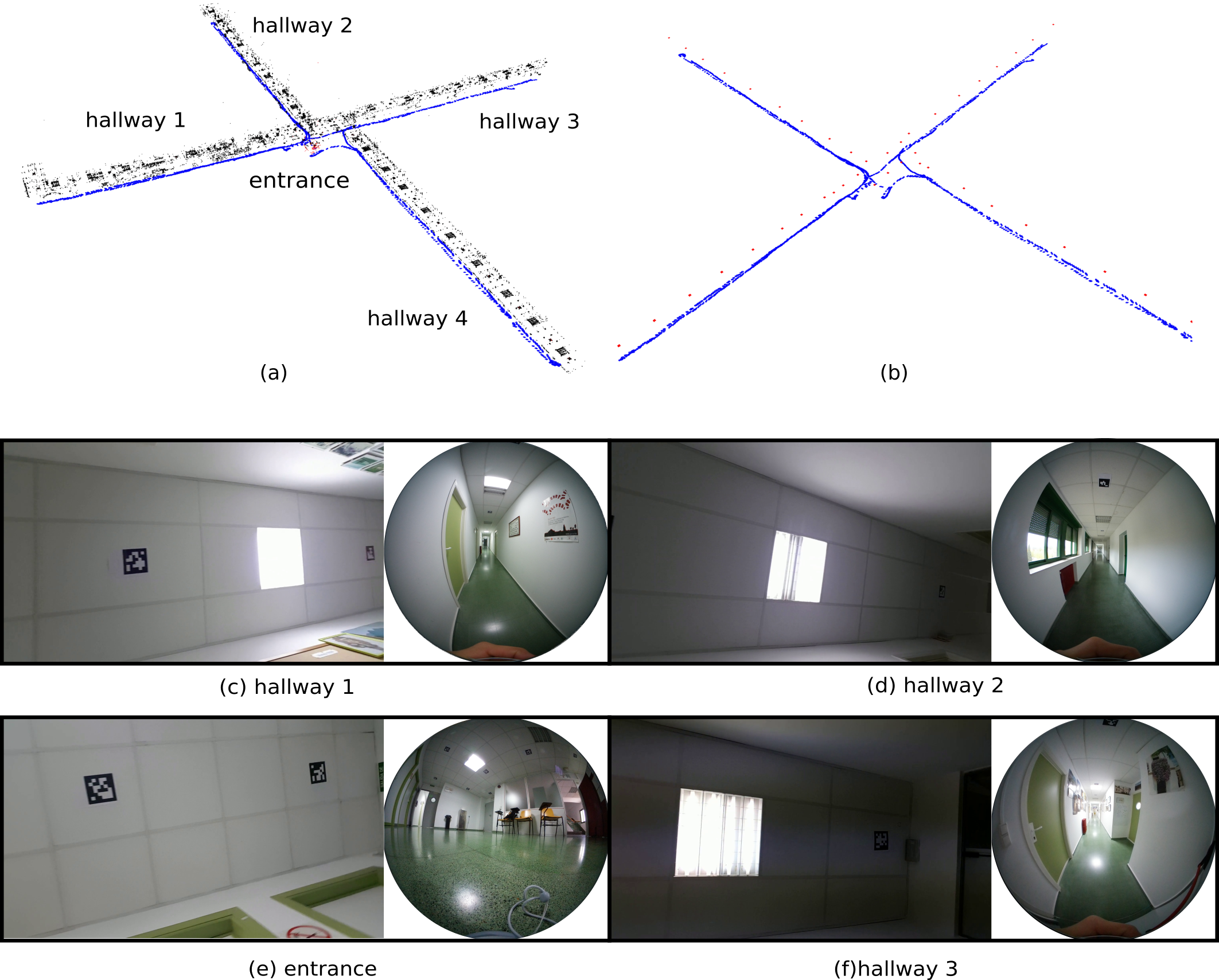}
\caption{ Large scale mapping in repetitive environment test. (a) reconstructed 3D view of the building floor. Black points are 3D map points. Blue elements are keyframes. The main regions are indicated in the image. (b) reconstructed 3D view without showing 3D points in order to visualize as red squares the fiducial markers detected. (c-d)  For each region, left image is a snapshot of the video sequence recorded to create the map while right image is a spherical view. Please notice that the views of the ceiling are very similar in all regions. It is impossible to solve the relocalization problem without using markers.}
\label{fig:ucoceiling_composed}
\end{figure*}

In this final experiment we present a special use case that can not be solved unless the proposed method is employed. We have placed a total of fifty markers on the ceiling of our building, which is comprised by a hall and four corridors of approximately twenty meters each. The average distance between markers is two meters. A video sequence  has been recorded with the frontal camera of a Samsung7 mobile phone, pointing to the ceiling while moving along the building. It has a total of $12.000$ frames, with a resolution of $1280\times720$ pixels. Some images from the video sequence are shown at the left of Figs~\ref{fig:ucoceiling_composed}(c-f). Additionally, we have presented at the right a fisheye view of the different parts of the building to better visualize the environment.

This type of scenario poses several challenges. First, beacuse of the lamps, the auto-gain system of the camera makes some images to appear with very low contrast (e.g. Figs~\ref{fig:ucoceiling_composed}(f)). Second, using exclusively keypoints for SLAM is very complicated: there is very little texture, and also, the images show very repetitive patterns, making relocalization using only keypoints problematic. Third, since the number of markers is limited, the previous work SPM \cite{spm-slam}, which requires at least two markers in each image in order to build a map, can not be used.

The recorded video sequence, has been processed using ORB-SLAM2, LDSO and our system in the three modes (only keypoints, only markers, and markers with keypoints). ORB-SLAM2 and LDSO were incapable of processing the whole sequence, getting lost after few frames. For our system, the first two modes where incapable to successfully process the video sequence.  When only keypoints are employed the system raised false loop-closure detections continously, making the system incapable of finishing the process. To avoid that problem, we repeated the experiments by disabling loop-closure from keypoints. Then, the problem is that because of relatively fast movements in some parts of the sequence, the image becames blurry and the detection of the keypoints can not be done. Then, the system get lost, enters in relocalization mode, and because the problem of repetitive patters, the relocalization fails.

The only method capable of creating a map of the environment was our system when both keypoints and markers were employed. However, to avoid false loop-closures, we disabled their detection using keypoints. So, loop-closure was exclusively based on markers. The results obtained by our method are shown in Fig.~ \ref{fig:ucoceiling_composed}(a,b). The blue elements represent map keyframes, the black dots represent map points, and markers are shown in red. Since spotting them in Fig.~\ref{fig:ucoceiling_composed}a is difficult because of the map points, Fig.~\ref{fig:ucoceiling_composed}b shows the generated map without black dots. We only can make a qualitative evaluation of the generated map since we do not have ground-truth for this sequence. However, a visual inspection indicates that map is a loyal representation of the environment.

\begin{figure*}[t!]
\centering
\includegraphics[width=1\textwidth]{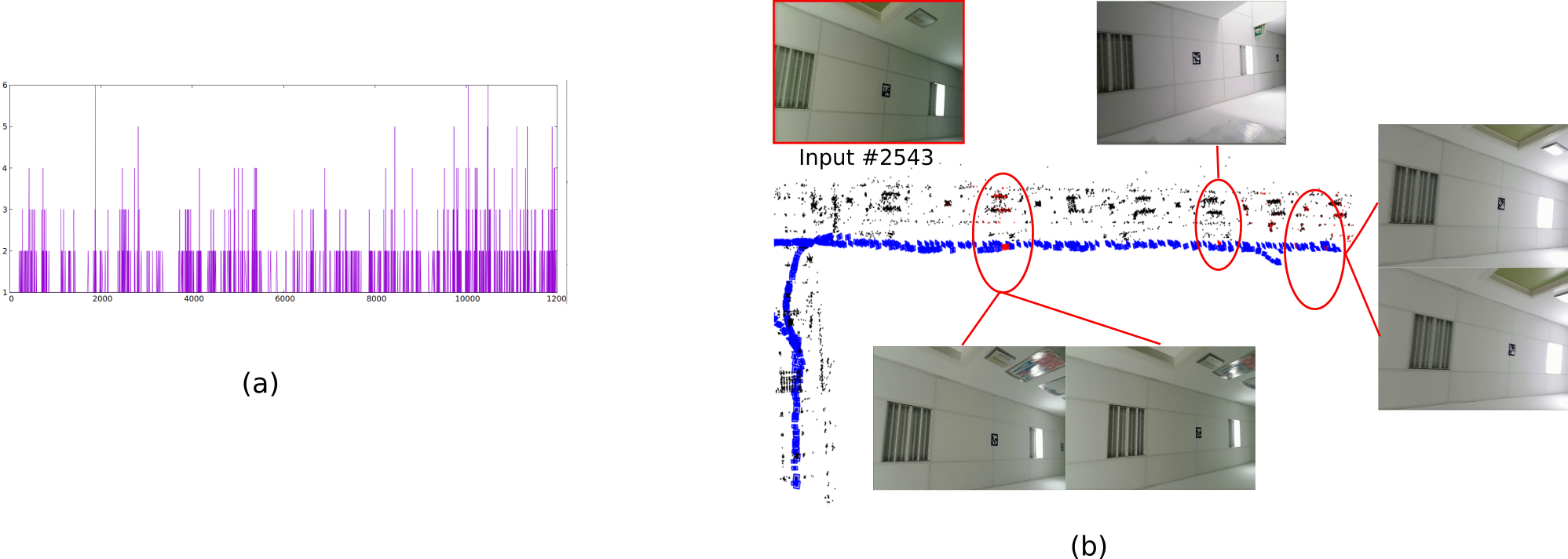}
\caption{Relocalization/Loop-Closure errors in repetitive environments. (a) Graph showing for each video frame, the number of incorrect relocalisation candidates found. (b) Erroneus relocalizations for frame \#2543 of the video sequence. See text for details.}
\label{fig:reloc_composed}
\end{figure*}

In order to show how repetitive the environment is, the following experiment has been conducted. Using the map generated, we have analyzed for each video frame  the number of false candidates for relocalization obtained using keypoints (the BoW database). The results are shown in Fig~\ref{fig:reloc_composed}(a), where the horizontal axis represents the video frame and the vertical axis the relocalization errors. As can be seen, many of the video frames have at least one incorrect relocalization keyframe. In some cases, there are up to six false relocalization candidates. To better understand the problem, Fig~\ref{fig:reloc_composed}(b) shows the case of frame \#2543. It has a total of five false relocalization candidates, that have been marked in the map along with the images corresponding to the keyframes. As can be observed, the input image is very similar to the false positives obtained. However, if the marker codes are analyzed, it is obvious that the problem is solved.

\subsection{Computing times}
\label{subsec::computingtimes}

In order to report the computing times of the proposed system, we   separate the map building task from the tracking task. Altough our system can be used as a real-time SLAM system, since it is executed in multiple threads,  we believe that a better approach is to create the map in a sequential mode, and then doing localization and tracking in the created map. In this way, no frames are dropped for the mapping.  Thus, we report separated computing the times for SLAM in sequential mode, and for tracking.

Table~\ref{tab:computingtimes_complete} shows the computing times of the different methods tested in Sect.~\ref{subsec::monoslam_test}. For each method, we report the average computing times expressed in frames per second. However, the values have been calculated considering the number framed successfully tracked. In other words, the values in the table are computed as the total time required to process the sequence divided by the total number of tracked frames. We proceed like this to avoid favouring a method when it has not estimated the pose of the camera. Plase consider that when a method get lost, the computing time of the lost frames becomes much smaller since they are not added to the map and no map optimization is required (which the most computationally demanding operation). The entries with value $0$ correspond to sequences for which a method was not able to perform initialization.
 
In the last row of the table, we present the average values obtained (without considering the $0$ values). As can be observed, the computing times of the proposed method are better than the other methods. In order to better understand the results, Wilcoxon Signed-Ranks have been performed to compare the methods. When our method is compared to LDSO, we observe no significant differences in the SLAM times. Please notice that the sequences with $0$ values where removed from the test. In Tracking times, the Wilcoxon tests indicates that the different observed are statically relevant using a significance level $0.01$. When compared to ORBSLAM2, the Wilcoxon test indicates that the observed differences are statically significant both for SLAM and Tracking. In other words, our method compares favourly in terms of computing time to the other methods and the differences are statically significant.

\section{Conclusions and future work}\label{sec:conclu}
 
This work has proposed a novel SLAM method combining keypoints and squared fiducial markers. The advantages of our method are several. First, our system can initialize both from markers and keypoints. Second, when markers are available, the real scale of the map is known. Third, the use of markers avoid false relocalizations and loop-closured in highly repetitive environments. Finally, markers can be used as long-term stable features in maps, in constrast to keypoints that can dramatically change over time.

In the experimentation conducted, it is shown that the proposed method performs better than state-of-the-art methods such as ORB-SLAM2 and LDSO  both in terms of precision and speed. For that comparison, we propose a novel measure to compare SLAM methods. Additionally, we show that the combined use of keypoints and makers improves accuracy and removes ambiguous situations in repetitive environments (such as offices or industrial facilities).

Finally, our implementation (which is publicly available \footnote{\url{ucoslam.com}}) allows to save the maps and reuse them later. A feature missing in other systems such as ORB-SLAM2.

\begin{table*}[p] 
\centering
    \small
    \begin{tabular}{m{2.0cm}||m{2.3cm}||m{.95cm}||m{1.1cm}|m{1.1cm}||m{1.1cm}|m{1.1cm}||m{1.1cm}|m{1.1cm}}
          
          \multirow{2}{1.0cm}{  Dataset }  &
          \multirow{2}{1.0cm}{  Sequence} &
          \multirow{2}{1.0cm}{  Cam} &
          \multicolumn{2}{m{1.5cm}||} {LDSO} & 
          \multicolumn{2}{m{1.5cm}||}{ORB-SLAM2} & 
          \multicolumn{2}{m{1.5cm}}{UcoSLAM}  \\
          
         \cline{4-9} &  & & SLAM & Tracking & SLAM & Tracking & SLAM & Tracking\\
       
       \hline \hline
Euroc-MAV  & V1\_01\_easy           & cam0 & $   0$ & $   0$ & $0.76$ & $13.5$ & $4.55$ & $22.7$\\
Euroc-MAV  & V1\_01\_easy           & cam1 & $   0$ & $   0$ & $0.74$ & $13.4$ & $4.51$ & $22.7$\\
Euroc-MAV  & V1\_02\_medium         & cam0 & $0.00$ & $   0$ & $0.56$ & $14.0$ & $0.63$ & $18.5$\\
Euroc-MAV  & V1\_02\_medium         & cam1 & $0.00$ & $0.20$ & $0.56$ & $14.0$ & $0.61$ & $18.3$\\
Euroc-MAV  & V1\_03\_difficult      & cam0 & $0.00$ & $   0$ & $1.03$ & $15.8$ & $0.67$ & $21.7$\\
Euroc-MAV  & V1\_03\_difficult      & cam1 & $0.00$ & $0.00$ & $0.83$ & $15.8$ & $0.98$ & $17.8$\\
Euroc-MAV  & V2\_01\_easy           & cam0 & $7.80$ & $7.34$ & $1.69$ & $14.6$ & $4.69$ & $24.6$\\
Euroc-MAV  & V2\_01\_easy           & cam1 & $3.32$ & $4.77$ & $1.55$ & $15.0$ & $4.61$ & $24.3$\\
Euroc-MAV  & V2\_03\_difficult      & cam0 & $   0$ & $   0$ & $1.13$ & $15.6$ & $0.66$ & $19.9$\\
Euroc-MAV  & V2\_03\_difficult      & cam1 & $4.09$ & $5.82$ & $0.98$ & $15.4$ & $0.76$ & $19.3$\\
Euroc-MAV  & mh\_01                & cam0 & $   0$ & $   0$ & $0.79$ & $12.8$ & $4.13$ & $23.3$\\
Euroc-MAV  & mh\_01                & cam1 & $5.60$ & $   0$ & $0.89$ & $13.4$ & $4.11$ & $22.4$\\
Euroc-MAV  & mh\_02                & cam0 & $5.77$ & $   0$ & $0.99$ & $13.1$ & $4.21$ & $22.8$\\
Euroc-MAV  & mh\_02                & cam1 & $5.78$ & $6.28$ & $1.01$ & $13.6$ & $4.70$ & $23.5$\\
Euroc-MAV  & mh\_03                & cam0 & $0.00$ & $   0$ & $0.60$ & $13.7$ & $4.04$ & $24.3$\\
Euroc-MAV  & mh\_03                & cam1 & $0.00$ & $   0$ & $0.58$ & $13.6$ & $4.04$ & $24.0$\\
Euroc-MAV  & mh\_04                & cam0 & $5.37$ & $4.50$ & $1.03$ & $14.3$ & $3.39$ & $25.0$\\
Euroc-MAV  & mh\_04                & cam1 & $6.24$ & $3.67$ & $1.12$ & $14.3$ & $3.40$ & $24.9$\\
Euroc-MAV  & mh\_05                & cam0 & $6.01$ & $   0$ & $1.02$ & $14.5$ & $3.95$ & $25.7$\\
Euroc-MAV  & mh\_05                & cam1 & $6.28$ & $   0$ & $1.03$ & $14.7$ & $3.58$ & $24.7$\\
Kitti      & 00                   & cam0 & $   0$ & $   0$ & $1.49$ & $14.0$ & $0.25$ & $19.7$\\
Kitti      & 00                   & cam1 & $2.57$ & $2.53$ & $1.46$ & $14.7$ & $0.25$ & $20.0$\\
Kitti      & 01                   & cam0 & $3.55$ & $3.22$ & $0.98$ & $12.6$ & $0.23$ & $20.0$\\
Kitti      & 01                   & cam1 & $3.70$ & $3.14$ & $0.55$ & $4.72$ & $0.18$ & $16.6$\\
Kitti      & 02                   & cam0 & $   0$ & $   0$ & $1.56$ & $13.9$ & $0.21$ & $19.2$\\
Kitti      & 02                   & cam1 & $0.72$ & $1.30$ & $1.63$ & $14.2$ & $0.21$ & $19.2$\\
Kitti      & 03                   & cam0 & $   0$ & $   0$ & $0.16$ & $0.29$ & $0.45$ & $18.2$\\
Kitti      & 03                   & cam1 & $   0$ & $   0$ & $1.28$ & $13.1$ & $0.27$ & $3.48$\\
Kitti      & 04                   & cam0 & $   0$ & $   0$ & $1.51$ & $13.5$ & $0.41$ & $19.3$\\
Kitti      & 04                   & cam1 & $   0$ & $   0$ & $1.50$ & $13.5$ & $0.45$ & $19.3$\\
Kitti      & 05                   & cam0 & $   0$ & $   0$ & $1.42$ & $13.1$ & $0.25$ & $18.9$\\
Kitti      & 05                   & cam1 & $   0$ & $   0$ & $1.44$ & $14.3$ & $0.29$ & $18.0$\\
Kitti      & 06                   & cam0 & $   0$ & $   0$ & $2.01$ & $12.3$ & $0.02$ & $20.7$\\
Kitti      & 06                   & cam1 & $   0$ & $   0$ & $1.73$ & $13.4$ & $0.02$ & $15.6$\\
Kitti      & 07                   & cam0 & $   0$ & $   0$ & $1.66$ & $14.9$ & $0.56$ & $19.3$\\
Kitti      & 07                   & cam1 & $   0$ & $   0$ & $1.83$ & $15.1$ & $0.58$ & $19.6$\\
Kitti      & 08                   & cam0 & $   0$ & $   0$ & $1.68$ & $14.1$ & $0.24$ & $19.0$\\
Kitti      & 08                   & cam1 & $   0$ & $   0$ & $1.71$ & $14.2$ & $0.24$ & $18.6$\\
Kitti      & 09                   & cam0 & $   0$ & $   0$ & $1.50$ & $11.5$ & $0.45$ & $20.1$\\
Kitti      & 09                   & cam1 & $   0$ & $   0$ & $1.89$ & $15.4$ & $0.46$ & $20.6$\\
SPM        & video1               & cam0 & $3.05$ & $2.33$ & $1.82$ & $4.16$ & $4.05$ & $12.8$\\
SPM        & video2               & cam0 & $2.09$ & $2.01$ & $2.23$ & $8.09$ & $4.74$ & $15.5$\\
SPM        & video3               & cam0 & $0.34$ & $0.56$ & $2.29$ & $8.90$ & $5.36$ & $17.0$\\
SPM        & video4               & cam0 & $0.57$ & $0.99$ & $2.92$ & $9.14$ & $6.24$ & $18.1$\\
SPM        & video5               & cam0 & $0.00$ & $   0$ & $1.92$ & $3.72$ & $6.18$ & $17.6$\\
SPM        & video6               & cam0 & $0.00$ & $   0$ & $0.53$ & $0.78$ & $5.76$ & $17.8$\\
SPM        & video7               & cam0 & $0.03$ & $0.25$ & $2.63$ & $6.92$ & $3.58$ & $11.7$\\
SPM        & video8               & cam0 & $   0$ & $   0$ & $0.03$ & $0.04$ & $3.89$ & $13.1$\\
TUM        & fr1\_desk2            & cam0 & $0.00$ & $   0$ & $0.15$ & $0.17$ & $2.31$ & $16.9$\\
TUM        & fr1\_desk             & cam0 & $7.55$ & $6.37$ & $0.95$ & $16.5$ & $3.82$ & $22.6$\\
TUM        & fr1\_floor            & cam0 & $5.21$ & $2.78$ & $2.86$ & $20.1$ & $5.22$ & $26.5$\\
TUM        & fr1\_xyz              & cam0 & $7.18$ & $8.13$ & $2.37$ & $19.9$ & $7.18$ & $22.1$\\
TUM        & fr2\_360\_kidnap       & cam0 & $0.00$ & $   0$ & $4.95$ & $11.8$ & $   0$ & $   0$\\
TUM        & fr2\_desk\_person      & cam0 & $0.00$ & $   0$ & $0.00$ & $0.00$ & $0.00$ & $0.02$\\
TUM        & fr2\_desk             & cam0 & $0.98$ & $1.88$ & $1.92$ & $17.3$ & $4.51$ & $24.0$\\
TUM        & fr2\_xyz              & cam0 & $8.39$ & $9.45$ & $7.92$ & $21.8$ & $7.56$ & $23.2$\\
TUM        & fr3\_long\_office      & cam0 & $6.54$ & $6.72$ & $1.65$ & $17.9$ & $3.62$ & $24.5$\\
TUM        & fr3\_nstr\_tex\_near    & cam0 & $3.98$ & $3.89$ & $5.56$ & $21.6$ & $6.05$ & $26.8$\\
\hline
Avrg       &                         &       &$\bf{3.0}$ & $\bf{2.4}$ & $\bf{1.6}$  & $\bf{12.5}$ & $\bf{2.6}$ & $\bf{19.8}$\\
\end{tabular}
    \caption{Speed  of the different methods (in frames per second) tested on the datasets Euroc-MAV~\cite{Burri25012016},  Kitti~\cite{Geiger2012CVPR}, SPM~\cite{spm-slam} and TUM~\cite{sturm12iros}. We report time required for SLAM and tracking only. }

    \label{tab:computingtimes_complete}
\end{table*}
\section*{Acknowledgment}
 This project has been funded under projects TIN2016-75279-P and IFI16/00033 (ISCIII) of Spain Ministry of Economy, Industry, and Competitiveness, and FEDER.

\bibliographystyle{elsarticle-num}
\bibliography{refs}

\end{document}